\documentclass{article}

\usepackage[a4paper,
            left=1.25in,
            right=1.25in,
            top=1.25in,
            bottom=1.25in,%
            footskip=.25in]{geometry}
            
\usepackage[colorlinks = true,
            linkcolor = blue,
            urlcolor  = blue,
            citecolor = blue,
            anchorcolor = blue]{hyperref}
            
\usepackage[numbers,sort,compress]{natbib}

\usepackage{pdfpages} 

\usepackage[labelfont=bf,labelsep=period]{caption} 

\usepackage{graphicx} 
\usepackage{amsmath,amssymb,mathtools}
\usepackage{makecell}
\usepackage{ulem} 
\title{\textbf{{\large Supporting Information for}}\\
Efficient automatic design of robots.}
\author{David Matthews, Andrew Spielberg, Daniela Rus, Sam Kriegman*, Josh Bongard}
\date{}

\begin{document}



# Efficient automatic design of robots

David Matthews[1], Andrew Spielberg[2], Daniela Rus[2], Sam Kriegman[1]*, Josh Bongard[3]

[1]Center for Robotics and Biosystems, Northwestern University, Evanston, IL 60208, USA. [2]Computer Science and Artificial Intelligence Laboratory, Massachusetts Institute of Technology, Cambridge, MA 02139, USA. [3]Department of Computer Science, University of Vermont, Burlington, VT 05405, USA.

*Corresponding author. Email: sam.kriegman@northwestern.edu

**Abstract:** Robots are notoriously difficult to design because of complex interdependencies between their physical structure, sensory and motor layouts, and behavior. Despite this, almost every detail of every robot built to date has been manually determined by a human designer after several months or years of iterative ideation, prototyping, and testing. Inspired by evolutionary design in nature, the automated design of robots using evolutionary algorithms has been attempted for two decades, but it too remains inefficient: days of supercomputing are required to design robots in simulation that, when manufactured, exhibit desired behavior. Here we show for the first time *de-novo* optimization of a robot's structure to exhibit a desired behavior, within seconds on a single consumer-grade computer, and the manufactured robot's retention of that behavior. Unlike other gradient-based robot design methods, this algorithm does not presuppose any particular anatomical form; starting instead from a randomly-generated apodous body plan, it consistently discovers legged locomotion, the most efficient known form of terrestrial movement. If combined with automated fabrication and scaled up to more challenging tasks, this advance promises near instantaneous design, manufacture, and deployment of unique and useful machines for medical, environmental, vehicular, and space-based tasks.

**Significance:** Artificial intelligence (AI) systems have taken massive strides designing prose, artwork, game play, software, and proteins, but have yet to master the design of complex physical machines. Here we introduce an automatic optimization method that can design self-moving machines — robots — from scratch by tracing failures in its behavior back to errors or inefficiencies in particular parts of its physical structure. Because this method improves the robot in this way, it can optimize the interdependent parts of the robot much more rapidly than the current approach, in which the designer tries different robot designs in a trial and error fashion. This opens the way toward bespoke AI-driven design of robots for a wide range of tasks, rapidly and on demand.

### Introduction

Autonomous robots pale in comparison to the complexity, adaptivity, and diversity of their living counterparts, animals, which emerged and radiated over billions of years of evolution. While

human minds can invent new mobile machines much faster than natural selection, the forms and functions of these artifacts are tightly constrained by human imagination as well as economics: considerable labor, time, money, and resources are required to design, test, and deploy a new robot for a new task. As a result, existing mobile robots tend to share canonical body shapes, such as quadrupedal or humanoid, canonical size scales, such as centimeters or meters in length, and canonical materials, such as metals and plastics. Despite this standardization, no proof exists that such robot body plans are the best fit for the robot's intended niche. If scalable design and deployment of robots were to be achieved, and those robots could be rendered recyclable (*1*) or biodegradable (*2*), robots of diverse size and shape could be deployed into millions of niches in natural and built environments, much like evolution and reproduction does in nature.

Thus, here we demonstrate an alternative, scalable approach for rapidly and cost-effectively designing new kinds of robot body plans with minimal prior assumptions: a custom robot was efficiently and automatically generated in less than 30 seconds on a laptop computer (Fig. 1), requiring only 10 design attempts. This is orders of magnitude more efficient than all reported robot design methods that utilized computer simulations (*1-10*) and an order of magnitude fewer design attempts than a method that forewent simulation (*11*). Although many different approaches to the automatic design of virtual robots have been reported in the literature (*12-30*), the automatic design of physical robots to date has been achieved predominately through evolutionary algorithms: design variants evolve through random variation and selection. The trial-and-error nature of this process can lead to thousands or even millions of robots being evaluated in virtual environments before selecting one suitable for physical fabrication (*1-10*). Other evolutionary methods have evolved physical robots directly, without recourse to simulation, but these require hundreds of physical trials, again reducing efficiency (*11*). Evolutionary approaches are appealing for robot design as, like biological evolution, they often discover diverse and non-obvious body plans and behaviors; they cannot, however, guarantee discovery of locally optimal designs, and they are inefficient as the majority of effort is spent evaluating evolutionary dead ends. For these reasons, almost every other aspect of automatic optimization has found success replacing such trial-and-error search with gradient descent methods, which can efficiently follow paths through design space to optima.

Applying gradient-based optimization to robot design has proven challenging because, like all motile organisms, there are complex feedback loops between the robot's body plan (e.g. its shape and distribution of motors) and its behavior. Determining which aspect of a robot's body plan caused some inefficiency in the robot's behavior, and how that body part could be changed to improve behavior, is a version of the credit assignment problem, a ubiquitous problem in artificial intelligence and, until now, unsolved for robot design.

The recent appearance of differentiable physical simulators (*31,32*) has enabled gradient-based design of virtual robots (*22,27,28*): aspects of the robot's shape or material properties responsible

for poor behavior can be identified and mitigated non-randomly to improve behavior. But the methods that produced them could not radically alter the robot's internal structure (musculature, mass distribution, and voids) or external structure, such as the addition of new limbs. Moreover, none of them were realized as physical robots. Other gradient-based approaches have been reported for optimizing the orientation or length of a robot's limbs (*33-37*) but the robot's overall structure (its distribution of limbs and motors) was presupposed and held fixed. Thus, here we introduce an algorithm that (i) simulates and then assesses the fitness of a virtual robot's behavior; (ii) identifies deficiencies in its overall shape, topology (number of voids), number and shape of limbs, mass distribution, musculature, and behavioral control; and (iii) changes all of them, simultaneously, so as to improve behavior in the next simulation. We demonstrate that only 9 repetitions of this process (and thus only 10 simulations) provides a design that, when built, retains the optimized function (Fig. 2). This is orders of magnitude more efficient than state-of-the-art algorithms for automated design of physical robots (*1-10*).

**Results**

Terrestrial locomotion was selected as the target behavior as it is the defining function of higher animals. The fitness of a simulated robot design was thus defined as its mean velocity in the desired direction of travel. Starting from a randomly-generated body plan, fitness was increased from negligible movement to 0.5 body lengths per second (BL/s) (half of average human walking speed) with just nine additional design iterations (Fig. 1). We then conducted another 99 trials of this process, starting from 99 different randomly-configured robots (Fig. S2) and observed statistically significant increases in velocity from 0 to a mean of 0.45 BL/s over the 10 simulations, across these 100 trials ($p < 0.01$; Fig. 3). The final robot design from the first trial was built (Fig. S1), and found to retain its behavior: when this physical optimized robot and another physical non-optimized robot were each evaluated several times, the former moved on average significantly faster than the latter ($p < 0.01$; Fig. 2I).

Several control experiments were performed to prove that the design process could and did make use of its freedom to change all aspects of the robot's shape, topology, and musculature to improve behavior, despite the complex interdependencies between them (Fig. 3A). In the first control experiment, an initial random placement of muscles was generated and held fixed. Only the robot's shape and topology was then optimized (Fig. 3B). In the second control experiment, muscle placement was optimized within a fixed, randomly-generated shape and topology (Fig. 3C). The first and second control experiments yielded simulated robots with significantly lower fitness than the final robots from the original experiment [0.31 (99%CI 0.30 to 0.33) and 0.08 (99%CI 0.07 to 0.08) BL/s, respectively, compared to 0.45 (99%CI 0.43 to 0.47) BL/s], proving that body plan and musculature were consistently co-optimized. This demonstrates that automatic optimization methods must be allowed to simultaneously track and correct inefficiencies across many different aspects of a robot's physical structure to achieve efficient

design of motile behaviors. The demonstrated interdependency between robot shape, musculature, and movement also shows that robot design optimization is not reducible to the more commonly explored problem of topology optimization for static structures (*38*).

The algorithm reported here dominates state-of-the-art automated design of physical robots, which is currently accomplished with gradient-free, evolutionary algorithms. Since all such methods (Fig. 1A) start with populations of more than 10 randomly-generated robots, they all reduce to random search when assigned a design budget of 10. We found that random search results in little to no appreciable forward movement (Fig. 3D) under these conditions.

Although our algorithm could have discovered crawling or peristaltic motion, it consistently rediscovered the third (and most efficient (*39*)) known form of terrestrial movement: legged locomotion (Fig. S3). It did so by moving or merging internal voids to create gaps between prospective limbs (Fig. 3E; Figs. S10-S13), and patterning muscles along each of their anterior edges to coordinate their movement (Fig. 3E). These changes consistently resulted in two or three posterior-angled appendages (Figs. 3F and S3). When the algorithm was forced to work around a randomly-generated and fixed musculature it failed to discover legged locomotion (Fig. 3B).

Fig. 4 reports the approach's generality: robots can be automatically and efficiently designed under other constraints, such as conservation of materials (Fig. 4A-D); for different tasks, such as object manipulation (Fig. 4E), object transport (Fig. 4F) and object ejection (Fig. 4G); and starting from different, arbitrary, non-rectangular initial shapes (Fig. 4H-L). Within these new tasks and experimental conditions, very different kinds of robot forms and functions emerged.

**Discussion**

While these results demonstrate efficient design automation for a mobile robot, much of the design pipeline remains to be optimized. For instance, the fabrication process includes several manual steps that could be automated with embedded 3D printing (*40*), the computational efficiency of the physics model could be optimized to run faster than realtime (*41*), and gradient computations could be optimized to be faster as well. Also, how the algorithm was allowed to alter the robot (moving and resizing internal voids or muscles) was somewhat arbitrary and may not readily extend to other kinds of materials (e.g. metals) and motors (e.g. rotary). Algorithms that are better at discovering and following gradients, and that generate robots from compressed representations (*42*) rather than directly rearranging their components, could likely be formulated. However, it is important to emphasize that gradient descent is inherently a local method and is therefore unlikely to find the globally optimal design — in the worst case, when designs are only of value if perfect, gradient descent will fail — but, in practice, gradient methods perform well in non-convex search spaces, find approximate global optima (*43*), and

can be re-deployed (Figs. S10-13) from different random initial configurations (Fig. S2). Finally, aspects of the physical robot's design were improved against just one feature (speed) of one behavior (locomotion); in nature diverse selection pressures simultaneously act on species. Thus, generalizing our approach such that alterations improve multiple features (reliability, safety, and energy efficiency) of multiple behaviors (object manipulation, collective behavior) remains unsolved but could in principle be achieved through multiobjective optimization (*44*) and optimizing the objective functions themselves (*45*). Despite these current limitations, the significant efficiency gains reported here compared to prior work suggests that this new approach can be more easily scaled to realize physical robots with more complex morphologies and behaviors, such as those already observed in simulation (Fig. 4).

Three decades ago, gradient-based optimization methods were introduced which allowed for efficient automatic design of neural networks (*46*). These networks are now affecting human health (*47*), the economy (*48*), and the environment (*49*), albeit indirectly via human or machine intermediaries. Scaling robotics by automating robot design could have as much, if not more impact, because robots could work directly within unique environments ranging from cells to asteroids. However this will only be possible if they have appropriate and tailored shapes, sizes, and material compositions to perform successfully and safely within them.

**Materials and Methods**

The design pipeline starts with a randomly-configured simulated robot and optimizes its structure over nine design iterations. The final robot design is then built and its physical behavior is assessed.

**Initial conditions.** Sixty-four voids and 64 muscle patches are placed randomly within a 20cm-by-14cm rectangular body. The initial location of each void was drawn from a uniform random distribution; the radius of each void was sampled from a gaussian distribution with a mean of 0.91cm and a standard deviation equal to the square of the mean. This corresponds to an initial void coverage (summed void areas; Sect. S2) of 60% of the robot's body area, on average. The initial location of each muscle patch was drawn from a uniform random distribution. The radius of each muscle patch was fixed at 1.26cm. This radius allows 64 nonoverlapping muscle patches to just cover the entire robot's body with musculature, in case such a distribution, or one similar to it, is useful.

The robot was modeled as a collection of elastic particles (Fig. 2) using the material point method (MPM) (*50*), a hybrid Eulerian-Lagrangian model (Sect. S1). Attempts to directly optimize the parameters of individual particles, without explicit voids, did not yield any changes in the robot's shape and resulted in significantly lower fitness (Fig. S5).

**Design.** Each design iteration comprises a forward and backward pass. During the forward pass, the robot is simulated using a differentiable simulator. At each time step of the simulation, internal forces from within the robot's body (its muscles and elastic deformation) and external forces from the environment (gravity, friction, collisions) are computed and the acceleration of each particle is updated. At the end of the evaluation period, a fitness score was automatically assigned to grade the robot's locomotive ability: the mean forward velocity of each particle during behavior, summed across particles. Since each function in this process is differentiable, the entire process, including the fitness computation, is differentiable.

During the backward pass, fitness affects the location and radius of each void (Eq. 1 in Sect. S1) and the location of each muscle (Eqs. 2 and 3 in Sect. S1). These functions are differentiable because they smoothly (quadratically) interpolate particle properties along the radius of an internal void or muscle, avoiding large discrete step functions at boundaries. More specifically, particle mass and elasticity decrease within a void, from edge to center, before entirely removing particles at their core (Figs. 2C and S7C). Muscles similarly interpolate volumetric actuation, increasing actuation amplitude from edge to center. When a void is moved or decreased in radius, particles are gradually added back into the robot's body (Sect. S1). Both musculature and voids were allowed to move off, and back on to, the body.

**Physical verification.** The physical robot is built by casting a silicone body with pneumatically actuated muscles (Sect. S3). To do so, two three-part molds of the design were 3D printed (Fig. S1). Each mold was then filled with silicone and allowed to cure to form one half of the robot's body. The two halves (left and right) of the design were sealed together with additional silicone and four retroreflective spheres were glued onto the top four corners to facilitate motion capture. Musculature was instantiated as a hollow bladder within the robot's body and was actuated pneumatically through a dorsal air inlet.

The robot's body and its environment were coated in cornstarch to match the friction parameters resident in the simulator. To compensate for asymmetries along the sagittal seam between the robot's two sealed parts, the robot was placed between two alignment rails (Fig. 1B). This helps ensure the robot's movement stays along a straight, rather than curved, path. However the rails are not strictly necessary (Movie S1).

The muscles were then actuated to create movement and allow for the possibility of locomotion in the desired direction. To do so, the robot's air inlet was connected to a proportional pressure control valve that generated a square wave with an amplitude of 300mBar, a wavelength of 500ms, and a duty cycle of 50%. The robot was actuated for 1 minute and its behavior was recorded with passive optical motion capture. After each behavior evaluation, the environment was reset: the robot was returned to its starting position, additional cornstarch was added to the terrain and raked to have a thin grooved surface. Six behavioral trajectories of the optimized

robot were collected and five were collected of the unoptimized robot (Fig. 2I). The optimized design moved significantly further on average than the unoptimized design ($p<0.01$; Mann-Whitney U).

**Generality.** Additional experiments were conducted *in silico* to better understand how (if at all) different random initial configurations (Fig. S2) affect the optimized body plan (Fig. S3), whether or not details of the manufacturing process could be incorporated as design constraints (Fig. S4), how different encodings alter the design space (Figs. S5 and S6), the algorithm's sensitivity to hyperparameters (Figs. S7-S9; Table S2), and the paths gradient descent followed through morphospace (Figs. S10-S13).

**Data Availability.** Source code is available in the GitHub repository (https://github.com/robodiff/robodiff). All other data are included in the manuscript and/or supporting information.

**List of Supplementary materials:** Movie S1, Methods Sects. S1-S4, Figs. S1-S13, Tables S1-S3.

**References**

1. Cellucci, D., et al. 1D printing of recyclable robots. *IEEE Robotics and Automation Letters*, **2**,1964-71 (2017)

2. Kriegman, S., et al. A scalable pipeline for designing reconfigurable organisms. *Proceedings of the National Academy of Sciences*, **117**, 1853-1859 (2020)

3. Lipson, H., Pollack, J. Automatic design and manufacture of robotic lifeforms. *Nature*, **406**, 974-978 (2000)

4. Hornby, G., et al. Generative representations for the automated design of modular physical robots. *IEEE Transactions on Robotics and Automation*, **19**, 703-719 (2003)

5. Hiller, J., Lipson, H. Automatic design and manufacture of soft robots. *IEEE Transactions on Robotics*, **28**, 457-466 (2011)

6. Moreno, R., Faiña, A. EMERGE modular robot: a tool for fast deployment of evolved robots. *Frontiers in Robotics and AI*, **198** (2021)

7. Kriegman, S., Scalable sim-to-real transfer of soft robot designs. *IEEE International Conference on Soft Robotics*, **3**, 359-366 (2020)

8. Kriegman, S., et al. Kinematic self-replication in reconfigurable organisms. *Proceedings of the National Academy of Sciences*, **118**, e2112672118 (2021)

9. Kriegman, S., et al. Scale invariant robot behavior with fractals. *Proceedings of Robotics: Science and Systems*, **17**, (2021)

10. Schaff, C., et al. Soft Robots Learn to Crawl: Jointly Optimizing Design and Control with Sim-to-Real Transfer. *Proceedings of Robotics: Science and Systems*, **18**, 10.15607/RSS.2022.XVIII.062 (2022)

11. Brodbeck, L., et al. Morphological evolution of physical robots through model-free phenotype development. *PloS One*, **10**, e0128444 (2015)

12. Sims, K. Evolving 3D Morphology and Behavior by Competition. *Artificial Life*, **1**, 353-372 (1994)

13. Bongard, J., Pfeifer, R. Repeated structure and dissociation of genotypic and phenotypic complexity in artificial ontogeny. *Proceedings of the Genetic and Evolutionary Computation Conference (GECCO)*, **3**, 829-836 (2001)

14. Hornby, G., Pollack, J. Evolving L-systems to generate virtual creatures. *Computers & Graphics*, **25**, 1041-1048 (2001)

15. Wampler, K., Popović, Z. Optimal gait and form for animal locomotion. ACM Transactions on Graphics (TOG), 28, 1-8 (2009)

16. Lehman, J., Stanley, K. Evolving a diversity of virtual creatures through novelty search and local competition. *Proceedings of the Genetic and Evolutionary Computation Conference (GECCO)*, **13**, 211-218 (2011)

17. Joachimczak, M., Wróbel, B. Co-evolution of morphology and control of soft-bodied multicellular animats. *Proceedings of the Genetic and Evolutionary Computation Conference (GECCO)*, **14**, 561-568 (2012)

18. Cheney, N., et al. Unshackling evolution: Evolving soft robots with multiple materials and a powerful generative encoding. *Proceedings of the Genetic and Evolutionary Computation Conference (GECCO)*, **15**, 167-174 (2013)

19. Lessin, D., et al. Open-ended behavioral complexity for evolved virtual creatures. *Proceedings of the Genetic and Evolutionary Computation Conference (GECCO)*, **15**, 335-342 (2013)

20. Geilinger, M., et al. Skaterbots: Optimization-based design and motion synthesis for robotic creatures with legs and wheels. *ACM Transactions on Graphics (TOG)*, **3**, 1-12 (2018)

21. Pathak, D., et al. Learning to control self-assembling morphologies: a study of generalization via modularity. *Advances in Neural Information Processing Systems (NeurIPS)*, **32**, (2019)

22. Spielberg, A., et al. Learning-in-the-loop optimization: End-to-end control and co-design of soft robots through learned deep latent representations. *Advances in Neural Information Processing Systems (NeurIPS)*, **32**, (2019)

23. van Diepen, M., Shea, K. A Spatial Grammar Method for the Computational Design Synthesis of Virtual Soft Locomotion Robots. *ASME Journal of Mechanical Design*, **141**, 101402 (2019)

24. Wang, T., et al. Neural Graph Evolution: Automatic Robot Design. *Proceedings of the International Conference on Learning Representations (ICLR)*, (2019)

25. Zhao, A., et al. Robogrammar: Graph grammar for terrain-optimized robot design. *ACM Transactions on Graphics (TOG)*, **39**, 1-16 (2020)

26. Gupta, A., et al. Embodied intelligence via learning and evolution. *Nature Communications*, **12**, 5721 (2021)

27. Ma, P., et al. DiffAqua: a differentiable computational design pipeline for soft underwater swimmers with shape interpolation. *ACM Transactions on Graphics (TOG)*, **40** (2021) 10.1145/3450626.3459832

28. Spielberg, A., et al. Co-learning of task and sensor placement for soft robotics. *IEEE Robotics and Automation Letters*, **6**, 1208-1215 (2021)

29. Yuan, Y., et al. Transform2Act: Learning a Transform-and-Control Policy for Efficient Agent Design. *Proceedings of the International Conference on Learning Representations (ICLR)*, (2022)

30. Hu, J., et al. Modular robot design optimization with generative adversarial networks. *Proceedings of the International Conference on Robotics and Automation (ICRA)*, 4282-4288 (2022)

31. Spielberg, A., et al. Advanced soft robot modeling in ChainQueen. *Robotica*, **41**, 74-104 (2023)

32. Hu, Y., et al. Taichi: a language for high-performance computation on spatially sparse data structures. *ACM Transactions on Graphics,* **38** (2019).

33. Sehoon, H., et al. Joint Optimization of Robot Design and Motion Parameters using the Implicit Function Theorem. *Robotics: Science and Systems* **8** (2017).

34. Spielberg, A., et al. Functional co-optimization of articulated robots. *In Proceedings of the International Conference on Robotics and Automation (ICRA)*, 5035-5042 (2017)

35. Du, T., et al. Computational multicopter design. *ACM Transactions on Graphics*, **35**, (2016)

36. Schulz A, et al. Interactive Robogami: An end-to-end system for design of robots with ground locomotion. *The International Journal of Robotics Research*, **36**, 1131-1147 (2017)

37. Chen, Desai, et al. Dynamics-aware numerical coarsening for fabrication design. *ACM Transactions on Graphics*, **36**, 1-15 (2017)

38. Sigmund, O., Maute, K. Topology optimization approaches. *Structural and Multidisciplinary Optimization*, **48**, 1031-1055 (2013)

39. Biewener, A. Biomechanics of mammalian terrestrial locomotion. *Science*, **250**, 1097-1103 (1990)

40. Truby, R., et al. Soft somatosensitive actuators via embedded 3D printing. *Advanced Materials*, **30**, 1706383 (2018)

41. Huang, W., et al. Dynamic simulation of articulated soft robots. *Nature Communications,* **11**, 2233 (2020)

42. Van Den Oord, A., et al. Neural discrete representation learning. Advances in Neural Information Processing Systems 30 (2017).

43. Ma, T. Why Do Local Methods Solve Nonconvex Problems? *Beyond the Worst-Case Analysis of Algorithms*, 465-485, Cambridge University Press (2021).

44. Fonseca, C., Fleming, P. An overview of evolutionary algorithms in multiobjective optimization. *Evolutionary Computation*, **3**, 1-16 (1995).

45. Houthooft, R., et al. "Evolved policy gradients." *Advances in Neural Information Processing Systems*, **31**, (2018)

46. Rumelhart D. et al. Learning representations by back-propagating errors. *Nature* **323**, 533–536 (1986)

47. Jumper, J., et al. Highly accurate protein structure prediction with AlphaFold. Nature 596, 583-589 (2021).

48. Zolas, N., et al. Advanced technologies adoption and use by US firms: Evidence from the annual business survey. No. w28290. National Bureau of Economic Research (2021)

49. Vinuesa, R., et al. The role of artificial intelligence in achieving the Sustainable Development Goals. Nature Communications 11, 1-10 (2020)

50. Sulsky, D., et al. A particle method for history-dependent materials. *Computer Methods in Applied Mechanics and Engineering*, **118**, 179-196 (1994)

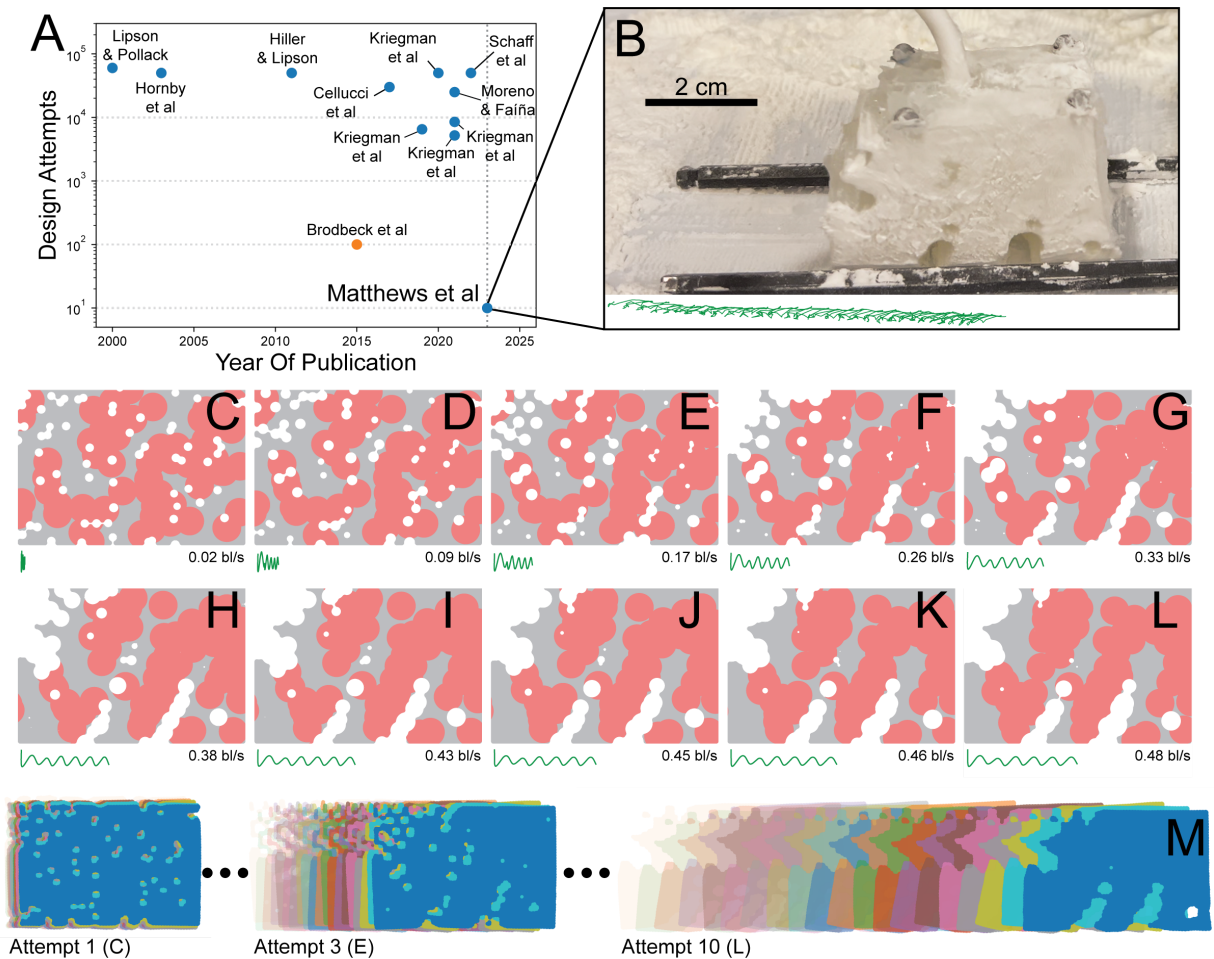

**Figure 1. Efficient automatic design.** (**A:**) The gradient-based approach introduced here takes 10 design attempts compared to thousands of simulations (blue dots in **A**; (*1-10*)) or hundreds of physical prototypes (orange dot in **A**; (*11*)) required by prior work. A mobile robot (**B**) was efficiently and automatically designed: starting from a randomly-generated morphology (gray body shape with red musculature; **C**), nine design revisions (**D-L**) improve locomotive ability (green CoM traces). **M:** The first, initially randomly-configured robot (C) moves up and down in place, while its nine descendents (attempts 2-10; D-L) exhibit increasingly better locomotive ability (movement into the right-hand side of the page).

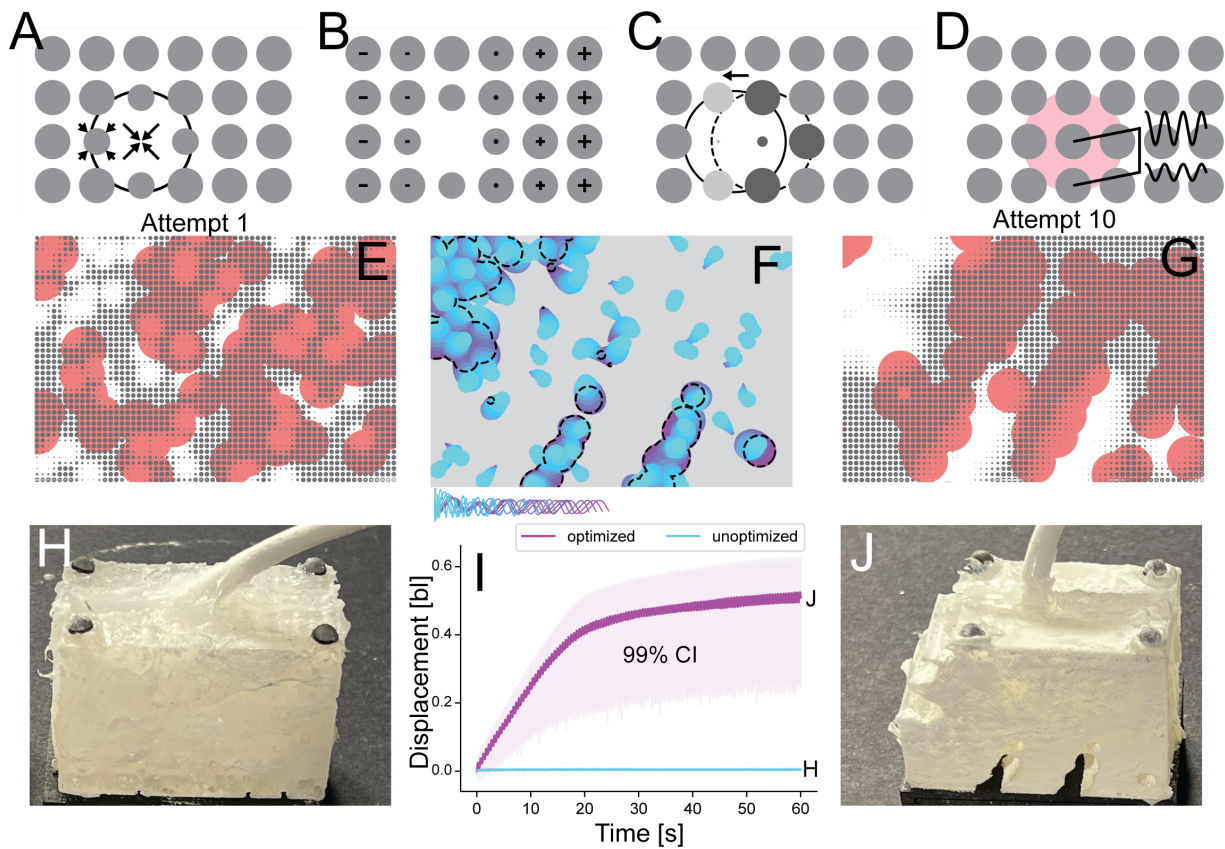

**Figure 2. Gradient-based optimization.** (**A:**) The robot was modeled as a two-dimensional plane of particles (gray), internal voids (circle; only one shown), and muscle patches (not shown). Voids reduce both the mass and elasticity of all particles within them. Particles closer to the center are reduced in mass/elasticity more; those very near the center are deleted entirely. (**B:**) During simulation, each particle's influence on the robot's behavior is tracked. After simulation, this information is used to calculate how each particle should be changed to improve the robot's behavior during the next simulation. In a simple example, such change is intuitive: if a robot incapable of moving should "move" to the right, the mass of particles to the left should be decreased (-) and those to the right should be increased (+). (**C:**) These desired mass/elasticity changes dictate void movement: particles that should have less and more mass/elasticity pull and push on nearby voids, respectively. Voids respond to these pressures by moving (arrow) and resizing (adjusting their radius; not shown), and then dictate the new masses/elasticities of all particles within them; previously-deleted particles may be recovered (dark gray). Using voids to indirectly affect particle masses/elasticities allows the design method to smoothly alter mass, elasticity, shape, and topology across the robot's body. (**D:**) Muscle patches (red) cause particles within them to volumetrically oscillate during simulation at the same phase, pushing and pulling against neighboring particles. As with voids, how the particles should have oscillated differently to improve behavior are calculated and used to move the muscle patches. (**E:**) Sixty-four movable and resizable voids and 64 movable muscles were randomly positioned within the

13

particle grid (Attempt 1), the resulting design was evaluated in simulation for locomotive ability, and the effect of each void and muscle on fitness was tracked. After simulation, how each void and muscle should be altered to improve fitness was calculated, and those alterations are made. The gradient of fitness with respect to each patch is followed for nine iterations (cyan to purple traces; **F**), resulting in an optimized design (Attempt 10; **G**). Once built, the randomly generated design (**H**) barely moves (**I**) whereas the optimized robot (**J**) exhibits legged locomotion.

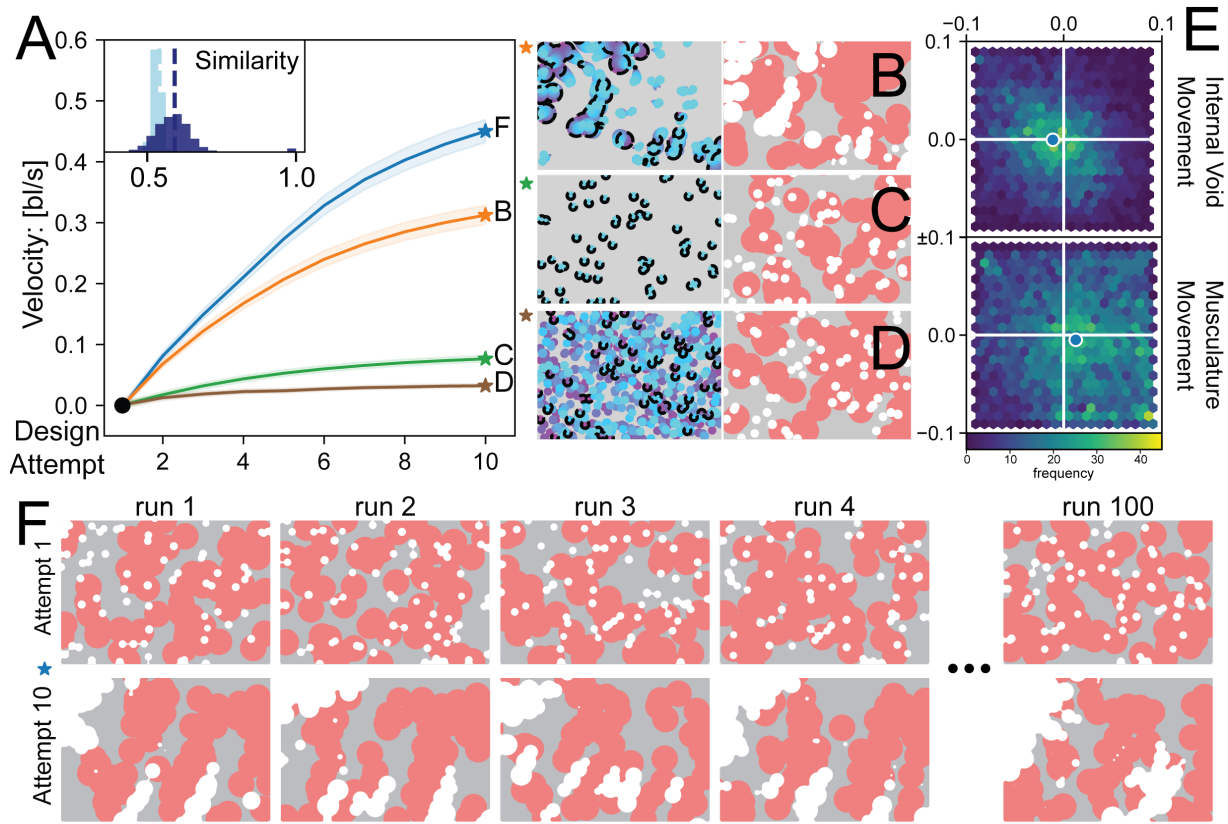

**Figure 3. Body shape and musculature.** To better understand the gradients leading to better designs and their sensitivity to where in design space optimization starts, ninety-nine additional independent runs of 10 design attempts each were performed, each starting from different random initial configurations. (**A:**) Following gradients with respect to both body shape and musculature, simultaneously (blue) resulted in significantly higher fitness (p<0.01) than optimizing body shape around a fixed musculature (orange; **B**) and optimizing musculature around a fixed body shape (green; **C**). Previous approaches to automate robot design were initialized with random samples of 10 or more designs; random samples of 10 designs yield little to no forward movement (purple; **D**). (**A**, inset:) The diversity of initial, random body plans before shape and musculature were co-optimized (light blue distribution) decreased significantly (p<0.01; Mann-Whitney U test) as the body plans converged on similar, legged designs when optimization completed (dark blue distribution). (**E:**) Internal voids tended either to move posteriorly (blue dot left of origin) or else shrink to zero radius and disappear, whereas muscles tended to move anteriorly (blue dot right of origin) as indicated by the heatmap of relative movement of muscles/voids from their initial locations (the origin). The result was tilted stacks of voids and muscles that form posterior-angled appendages with anterior musculature (**F**).

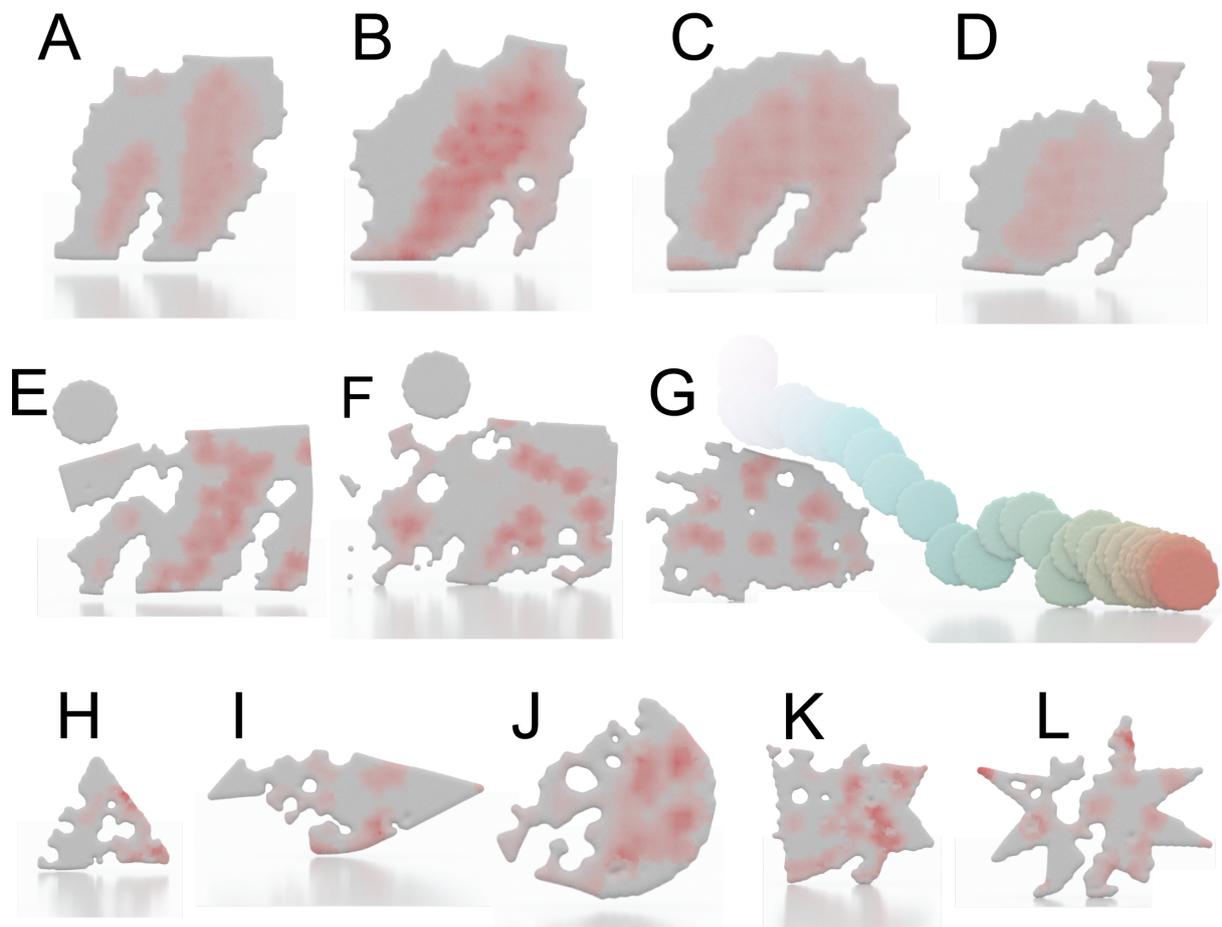

**Figure 4. Generality of results.** The algorithm introduced here can be extended to design problems that require efficient use of material (e.g. maximizing speed while minimizing surface area; **A-D**); tasks other than locomotion, such as object manipulation (**E**), transport (**F**), and ejection (**G**); and may also begin from arbitrary, non-rectangular, initial boundary conditions such as triangles (**H,I**), a circle (**J**), a heptagram (**K**) or enneagram (nine-sided star; **L**).



\maketitle

*To whom correspondence should be addressed.\\
Email: 
\href{mailto:sam.kriegman@northwestern.edu}{sam.kriegman@northwestern.edu} \\ \\

\textbf{This PDF file includes:}

Supporting text \\
SI References \\ 
Figures S1 to S13 \\
Tables S1 to S3 \\
Legend for Movie S1 \\

\textbf{Other supporting materials for this manuscript include the following:} 

Movie S1\\
Software S1 \\

\tableofcontents



\section*{Legend for Movie S1:}
\addcontentsline{toc}{section}{\protect\numberline{}Legend for Movie S1.}%

\href{https://youtu.be/qe_4_9cDt4Y}{\textbf{This movie}} summarizes the paper's results (a novel robot is designed from scratch in just 10 design attempts) and methods (using gradient based optimization).

\refstepcounter{section}
\section*{S1. Simulation.}
\label{sec:simulation}
\addcontentsline{toc}{section}{\protect\numberline{}Sect. S1: Simulation.}%

Each design attempt was modeled by a 64-by-44 partially-filled uniform grid of present/absent elastic particles (at most 2816 particles; 20-by-14 cm$^2$) 
and 
simulated 
using the Material Point Method (MPM; \cite{Sulsky1994}) 
for $T=1024$ timesteps 
with step size $dt=0.001$ sec, 
for a total evaluation period of 1.024 \textit{simulated} seconds.
One second of simulation time can be computed faster or slower than realtime, depending on the step size.
Smaller step sizes are more accurate than larger step sizes but require more steps (and thus more wallclock time) to simulate 1 sec. 
The number of timesteps is also constrained by memory and the accumulation of floating point rounding errors.
\\

In the main experiment, 
which culminated in the fabrication and testing of an automatically-designed physical robot (Figs. 1 and 2), 
and was repeated 99 times from different random initial configurations (Fig.~3),
a design consists of at most 64 internal voids, 
which remove contiguous circular regions of particles from the body grid, 
and at most 64 muscles, 
which expand and contract in volume by increasing and decreasing the distance between their constituent particles.
The 64 voids are moved and resized by optimization; a
the 64 muscles are moved by optimization but have have fixed size (radius of actuation, $r_c=1.26$ cm).
Voids and muscles can be removed based on their location (off the grid);
voids can also be removed by reducing their radius to 0.
Unless otherwise noted, we will assume the hyperparameters used by the main experiment.
\\

The mass and elasticity of each particle was based on its position relative to nearby voids.
For each particle, $\varphi \in \Phi$, and void, $k\in K$, the euclidean distance, $d_{\varphi,k}$, was calculated between every particle-void pair,
$\{\varphi,k\}$, and divided by the void's current radius, $r_k$, yielding a normalized distance, $d_{\varphi,k}^* = \min(1, d_{\varphi,k} / r_k)$.
The mass of particle $\varphi$ was then computed as:
\begin{equation}
\label{eq:mass}
m_{\varphi} = \left[\, \min_{k \in K}(d_{\varphi,k}^*)\right]^2 \times \text{1 kg} \, ,
\end{equation}
the square of the minimum normalized distance from the particle to every void in the body.
Particles toward the center of a void, with a mass less than $\lambda = 0.1$ kg, were removed from simulation.
The particle's elasticity, $E_{\varphi}$, was set to be proportional to its mass: $E_{\varphi} = m_{\varphi} \times \text{20 Pa}$.
\\

Simulated muscles undergo six actuation cycles during the 1024 timesteps of simulation.
At each time step, viscous damping is applied to each particle, the actuation state of each particle is calculated, and a vertical expansion or contraction force is generated.
The actuation amplitude of each particle is based on its position relative to nearby muscles.
For every muscle, $c \in C$, the euclidean distance,
$d_{\varphi,c}$,
between each particle-muscle pair,
$\{\varphi,c\}$,
was calculated and normalized as in Eq.~\ref{eq:mass}.
The actuation amplitude of particle $p$ was then computed as:
\begin{equation}
    \label{eq:actuation}
    A_{\varphi} = \left( 1 - \left( d_{\varphi,c}^* \times \sqrt{0.1} \right) \right) ^2,
\end{equation}
a quadratic interpolation with a minimum amplitude of 0.45 at the border of each muscle patch.
Particles outside of muscle patches, with $d_{\varphi,c}^* = 1$, were assigned zero actuation amplitude.
\\

The actuation state of each particle at simulation time $t$ was:
\begin{equation}
\label{eq:actuationProfile}
    \omega_{\varphi,t} = 4 \times m_{\varphi} \times \tanh(A_{\varphi} \times \sin(t \times dt \times f)),
\end{equation}
the hyperbolic tangent of a sinusoidal function with amplitude $A_{\varphi}$ and frequency $f=\text{6 Hz}$, 
normalized by the particle's mass and relative to a global actuation strength of 4 Pa per particle.
Normalizing by mass helps avoid numerical instabilities caused by tearing, which can occur from high particle acceleration due to low mass and high force.
\\

Other
void
(Fig.~\ref{fig:S8RobustnessToHyperparameters}A)
and muscle 
(Fig.~\ref{fig:S8RobustnessToHyperparameters}B)
radii,
other numbers of 
voids 
(Fig.~\ref{fig:S8RobustnessToHyperparameters}C)
and muscles 
(Fig.~\ref{fig:S8RobustnessToHyperparameters}D),
other interpolation functions for the particles inside voids
(Fig.~\ref{fig:S8RobustnessToHyperparameters}E) 
and muscles
(Fig.~\ref{fig:S8RobustnessToHyperparameters}F),
and 
other thresholds for particle removal within voids
(Fig.~\ref{fig:S8RobustnessToHyperparameters}G)
were explored in a subsequent hyperparameter sweep 
(Figs. \ref{fig:S7RobustnessToHyperParametersKey}-\ref{fig:S9RobustnessToHyperparametersGallery})
and detailed in Table~\ref{table:design}.
\\

\refstepcounter{section}
\section*{S2. Optimization.}
\label{sec:optimization}
\addcontentsline{toc}{section}{\protect\numberline{}Sect. S2: Optimization.}%

\textit{Initialization}. 
Optimization begins with a randomly generated design. 
Random designs comprised 64 internal voids and 64 muscles, randomly sized and positioned along the robot’s body. 
The position of each patch was selected from a uniform distribution across the robot’s body (20-by-14 cm$^2$). 
The radius of each muscle was fixed at $r_c=1.26$cm. 
The initial radius of each void was drawn 
from a normal distribution 
with mean $\mu=0.046 \times 20$cm 
and standard deviation of $0.046^2 \times 20 = 0.042$cm.
This removes
60\% of the 2816 particles, 
on average,
under the assumption that the voids are non-overlapping.
In other words, 
if
the 64 sampled voids 
are arranged side-by-side on a grid,
their a combined area
(their ``grid coverage''; $O_K$)
will be approximately $O_K=60\%$
of the robot's body area, on average.
Several other distributions of initial radius (Table~\ref{table:design})
were explored
in the hyperparameter sweep (Fig.~\ref{fig:S8RobustnessToHyperparameters}A,B).
\\

\textit{Fitness}. The fitness of a design was taken to be the mean forward velocity of the robot, which was calculated as the mean position of all of the particles at the final time step subtracted by the mean position of all of the particles at the first time step.
\\

\textit{Loss}. Gradient descent toolkits typically default to minimizing an error function rather than maximizing a fitness function.%
\footnote{
Gradient based optimization methods have historically used the metaphor of a ball on a hill: descending a slope (of error or ``loss'') within a landscape defined by some objective function. 
Evolutionary biology has instead used the metaphor of hill climbing: evolution as an active force that ascends slopes along a fitness landscape. 
These two metaphors, of maximizing fitness (negative error) and minimizing error (negative fitness), are functionally isomorphic.
}
For this reason, the loss function was defined as -1 $\times$ average forward velocity. That is, negative fitness (locomotive deficiency). 
\\

\textit{Backpropagation}. Backpropagation was performed \cite{Griewank1989} through the differentiable simulator, yielding gradients of loss (negative fitness) with respect to initial particle parameters. Initial attempts to directly optimize these initial particle parameters did not yield any changes in topology (Fig.~\ref{fig:S5DirectParticle}). 
Internal voids, if they possess non-zero radius, explicitly remove particles, and thus enable robot topology optimization. 
Gradients were therefore backpropagated one step further, from the particle parameters through the patch-based design functions (Eqs.~\ref{eq:mass}-\ref{eq:actuationProfile}), 
yielding gradients of loss with respect to the patch parameters (positions and sizes of voids, and positions of muscles). 
\\

\textit{Gradient descent}. Gradient descent using the Adam optimizer \cite{Kingma2014} with a learning rate of 0.01 was used to optimize the design parameters. Gradients from simulation were used to perform nine gradient descent steps, which, including the initial randomly generated design, was 10 design attempts.

\refstepcounter{section}
\section*{S3. Physical experiments.}
\label{sec:physicalExperiments}
\addcontentsline{toc}{section}{\protect\numberline{}Sect. S3: Physical Experiments.}%

\textit{Simulation-to-reality transfer}. 
The physical robot (5 cm wide) was manufactured as a 4:1 scale model of its simulated counterpart (20 cm wide), and driven by pneumatic actuation. 
To do so, the soft fringe of quadratically decreasing mass/elasticity
inside each simulated void  (Eq.~\ref{eq:mass}) was sharpened with a binary step function describing where silicone will and will not be present. 
The particles closest to the fully-removed core of each simulated void, 
which have very low stiffness and minimally contributed to robot's structure, were removed before building the design.
More specifically, 
the physical voids fully remove material in a radius that is 1.3 times larger than
the fully removed portion of the simulated voids.
Outside this radius, material is fully present. 
The physical void radius is also used in the figures when displaying void area rather than their influence on each particle.
\\

In simulation, muscles expanded along the y axis and were permitted to ``hang off the body'' along the exterior surface of the design. 
When manufacturing the robot as a silicone body with hollow bladders, actuated regions within 3 mm of the edge of the body were filled with passive material (Fig.~\ref{fig:S4ConstrainedActuation}A). 
Incorporating this post-processing step as a optimization constraint
(Sect.~\ref{sec:ConstrainedActuation}) results in the very same locally-optimal design when optimizing from the same randomly generated initial design.
\\

\begin{figure}[!h]
    \centering
    \includegraphics[width=\linewidth]{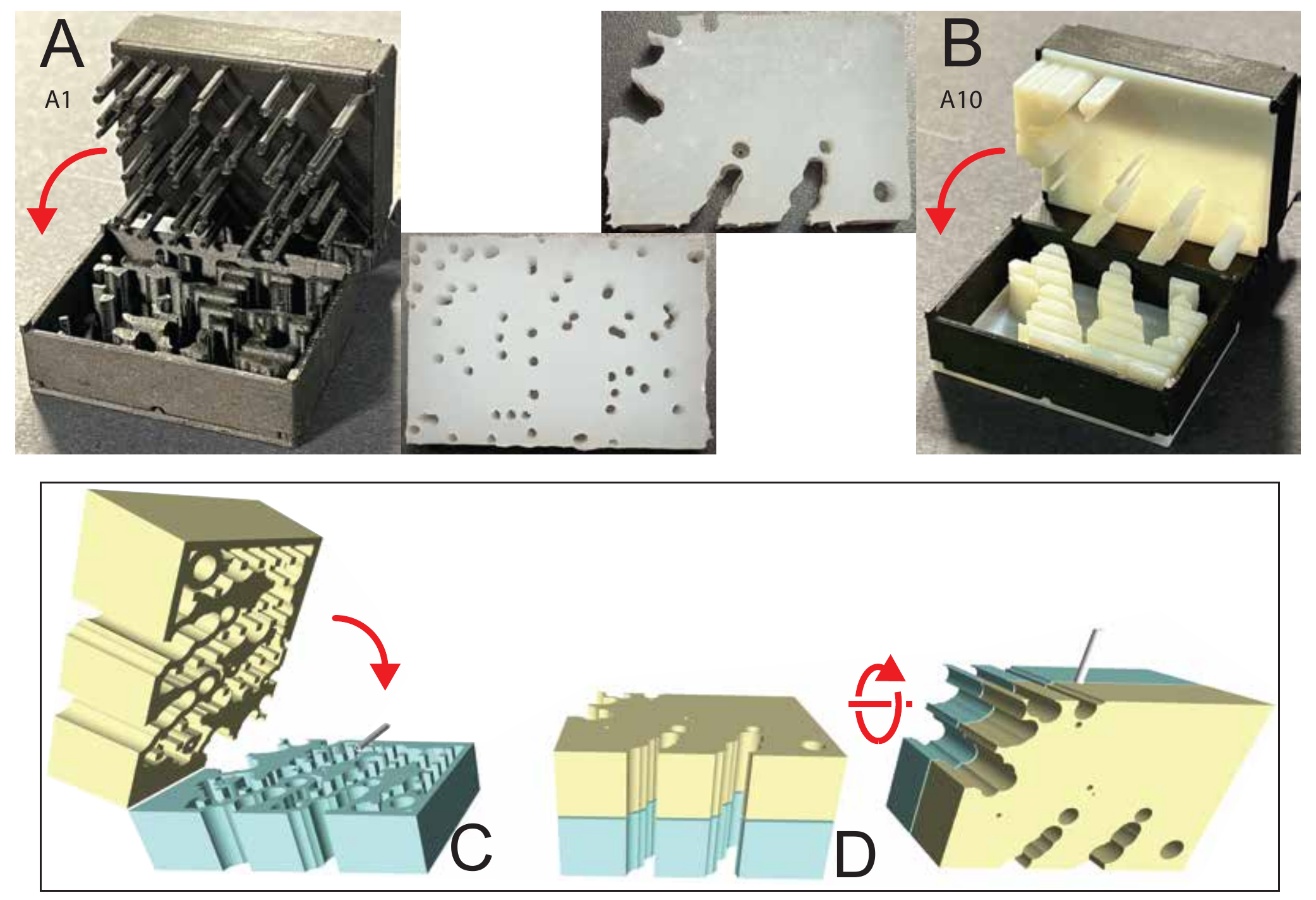}
    \caption{\textbf{Manufacture}. A set of molds were 3D printed for each of the initial (\textbf{A}) and final (\textbf{B}) design attempts (mirrored molds of a-b not shown). Silicone was poured into the molds, and two robot halves (bottom: yellow, blue) were removed. A pneumatic tube was glued into the blue mold with Sil-Poxy and then the yellow robot half was siliconed onto the blue robot half and allowed to cure (\textbf{C}). After curing, the robot was rotated into an upright posture and connected to a pneumatic pressure control system (\textbf{D}).}
    \label{fig:S1Manufacturing}
\end{figure}

Physical muscles were connected together with small cutouts, and horizontal stiffeners were added to the actuation bladders to limit the horizontal component of expansion when pressure is applied. 
Finally, a pneumatic port was also added to the center of the top of the robot.
The robot was split along the sagittal plane, and a three-part mold was generated for each half (Fig.~\ref{fig:S1Manufacturing}). 
The mold (six parts in total) was then printed using a Markforged X7 3D printer with Onyx filament.
\\

\textit{Casting the robot}. 
Each mold part was covered in a mold release. 
The bottom and middle parts of the mold were assembled (Fig.~\ref{fig:S1Manufacturing}A-B). 
Smooth-On Dragon Skin 10 Fast was mixed and degassed in a vacuum chamber for 7 min. Each mold was filled with silicone, and the third mold piece (lid) was inserted. 
A weight was then placed on top of the lid to hold it in place during the 75 min curing period. 
The cured robot halves were removed from the molds, and silicone flashing was trimmed off with scissors and a scalpel. 
Silicone rubber tubing was glued into one of the robot halves with Sil-Poxy and allowed to cure for 12 min. 
The two robot halves were then siliconed together with another batch of degassed silicone and allowed to cure for another 75 min. 
Four retroreflective balls were glued onto the top four corners of each robot using Sil-Poxy.
\\

\textit{Evaluation}. 
Cornstarch (Argo) was placed on top of a sheet of plastic and raked to have a thin grooved surface. 
Two alignment rails were added to the surface to simplify the comparison 
of the simulated robot (which can only move in a straight line)
and 
the physical robot (which has a slightly curved gait due to asymmetries along its sagittal seam where its two halves meet).
The robot was placed between the two alignment rails and connected to a Festo proportional pressure control valve (PPM-6TA-L-1-F-0L2H-S1) that generated a square wave with an amplitude of 300mBar, a wavelength of 500ms, and a duty cycle of 50\%. 
The robot was actuated for 1 min and its behavior was captured using a motion capture system (OptiTrack).
After evaluating the robot's behavior, the environment was reset.
Six behavioral trajectories of the optimized robot (Attempt 10) were collected and five were collected of the unoptimized robot (randomly generated; Attempt 1).
The optimized design moved significantly further on average (winning design vs randomly generated robot: MannWhitneyU test; $p < 0.01$).

\refstepcounter{section}
\section*{S4. Supplemental experiments.}
\label{sec:supplementalExperiments}
\addcontentsline{toc}{section}{\protect\numberline{}Sect. S4: Supplemental Experiments.}%

Additional experiments were conducted \textit{in silico} to 
promote the efficient use of building material (Sects.~\ref{sec:MatErosion} and \ref{sec:RotationalMoment}),
round sharp edges and corners on the body
(Sect.~\ref{sec:CircleMask}),
re-place deactivated voids/muscles
(Sect.~\ref{sec:PatchReplacement}),
include agnostic and antagonistic musculature
(Sect.~\ref{sec:sinCosAct}),
achieve other behaviors besides locomotion
(Sect.~\ref{sec:ObjectManipulation}),
enable interactive design from user-defined initial morphologies
(Sect.~\ref{sec:MaskedDesigns}),
incorporate details of the manufacturing process as optimization constraints instead of post-processing steps
(Sect.~\ref{sec:ConstrainedActuation}),
directly optimize particles instead of explicit voids and muscle patches
(Sect.~\ref{sec:DirectParticleOptimize}),
and
explore
other methods of interpolation within voids (Sect.~\ref{sec:VoidInterpolation})
and muscles (Sect.~\ref{sec:MuscleInterpolations}),
other thresholds for particle removal within voids
(Sect.~\ref{sec:ThresholdForRemoval})
other initial distributions of void (Sect.~\ref{sec:VoidRadiiDistributions})
and muscle (Sect.~\ref{sec:MuscleRadiiDistributions}) radius,
as well as different numbers of voids (Sect.~\ref{sec:NumberOfVoids})
and muscles (Sect.~\ref{sec:NumberOfMuscles}).
We repeated each experiment (each \textit{condition}) across several independent trials; each repeated trial starts from a unique random initial \textit{configuration} (a unique random seed).

\refstepcounter{subsection}
\subsection*{4.1. Eroding unnecessary body parts (Fig.~4A-B,E-G).}
\label{sec:MatErosion}

To promote efficient use of building materials
we combined
the gradients from simulation
with 
a global material erosion term,
which encourages the removal of unnecessary body parts by applying a pressure for a robot's total mass to reach some target amount.
More specifically, the gradient of fitness (loss; $\ell$) 
with respect to particle mass ($m_{\varphi}$) 
was computed as a linear combination of simulation loss ($\ell_{\text{sim}})$ and the erosion term:
\begin{equation}
    \frac{\partial \ell}{\partial m_{\varphi}}
    = 
    \frac{\partial \ell_{\text{sim}}}{\partial m_{\varphi}} 
    +
    \alpha \times
    \left[\, 
    \max_{\varphi \in \Phi} 
    \left(
    \frac{\partial \ell}{\partial m_{\varphi}}
    \right) 
    - 
    \min_{\varphi \in \Phi} 
    \left(
    \frac{\partial \ell}{\partial m_{\varphi}} \right)
    \right]
    \times 
    \tanh
    \left(
    \beta \times 
    \left(\bar{m}_{\varphi} - m^{\star} \right) 
    \right),
\end{equation}
where 
$\bar{m}_{\varphi}$
is the
average particle mass (including removed particles with zero mass);
coefficients
$\alpha = 0.01$
and
$\beta = 10$;
and
a target average mass
$m^{\star} \in \{0.25, 0.5, 0.75\}$.
The hyperbolic tangent function
is an S-shaped curve that
caps the material erosion gradients at 1,
and
the constant $\beta$ 
compresses the curve to be steeper and more rapidly approach 1.
Thus,
with $\alpha=0.01$, 
the magnitude of the erosion term 
is scaled to 1\% of the range of particle gradients from simulation.
As a result,
gradient descent will tend only to 
remove (or add) material 
in ways that minimally impact 
the behavior of the robot.
Pressure to add material occurs if the robot's mass falls below the target;
however, the target mass was chosen to be low enough that this was not observed.
\\

\refstepcounter{subsection}
\subsection*{4.2. Minimizing rotational moment (Fig. 4A-B,F).}
\label{sec:RotationalMoment}
In addition to augmenting the gradients through a material erosion term (Sect. \ref{sec:MatErosion}), a second augmentation term to minimize rotational moment of inertia was also introduced.
This term helps to keep the robots compact, especially when used in combination with other terms which encourage removal of material writ large (Fig. 4A-B, F).
To do so we first computed the robot's center of mass, $\Phi_{x,y}$,
as well as the euclidean distances
between 
$\Phi_{x,y}$
and
each particle $\varphi$, which we denote $\nu_{\varphi}$.
The rotational moment was then computed as:
\begin{align}
    \nu_{\varphi} &= D(\varphi, \Phi_{x,y}) \\
    h(\nu_{\varphi}) &= \begin{cases}
        \nu_{\varphi} & \text{if } \nu_{\varphi} < 0.25 \\
        2\nu_{\varphi} - 0.25 & \text{otherwise}\\
    \end{cases} \\
    \ell_I &= \sum_{\varphi \in \Phi} (m_{\varphi} \times h( \nu_{\varphi} )).
\end{align}
This loss was then backpropagated through to the mass of each particle and linearly combined with the particle gradients from simulation:

\begin{equation}
    \frac{\partial \ell}{\partial m_{\varphi}} = \frac{\partial \ell_{\text{sim}}}{\partial m_{\varphi}} + \gamma \frac{\ell_I}{\partial m_{\varphi}},
\end{equation}
where $\gamma \in \{ 0.0001, 0.0005, 0.001\}$.
\\

\refstepcounter{subsection}
\subsection*{4.3. De-brickifying bodies (Fig. 4C-D).}
\label{sec:CircleMask}

Initializing designs on a rectangular grid 
left a bias for brick-like bodies 
with 
square
dorsal-anterior 
and
ventral-posterior
edges.
To mitigate this bias we explored an alternative fitness (loss) function which rounds the edges of the body:
\begin{align}
    \ell_{\circ}
    &=
    \sum_{\varphi \in \Phi}
        m_{\varphi}
        \left(
        2 
        \left(
            d_{\varphi, \circ} > r_{\circ}
            \right) - 1
             \right) \\
    \ell &= \ell_{sim} + \gamma_{\circ} \times \ell_{\circ},
\end{align}
where $\gamma_{\circ} \in [0.00001, 0.0001]$.
This additional term promotes 
the addition of material inside of a circle \raisebox{-.2ex}{\Large$\circ$} of radius $r_{\circ}$, 
and the removal material outside of it.
\\


\refstepcounter{subsection}
\subsection*{4.4. Replacing deactivated patches (Fig. 4A-G).}
\label{sec:PatchReplacement}

When a patch (void or muscle) is 
removed
(reduced to radius zero,
moved off the body,
or overlapping another patch)
it no longer impacts 
particle mass (in the case of voids) nor 
actuation (in the case of muscles)
and 
thus no longer
receives any gradients from simulation.
It remains to determine if deactivating patches in this way 
caused or could cause
premature convergence to suboptimal designs.
If it does, a potential solution could be 
re-placing each removed patch
back onto the body at a new, random location,
and restoring its initial radius. 
This strategy was explored
for the robots presented in Fig. 4A-G.
\\


\refstepcounter{subsection}
\subsection*{4.5. Agonistic and antagonistic musculature (Fig.~4H-L).}
\label{sec:sinCosAct}
Agonistic and antagonistic musculature
were implemented 
using 
two
different
sets of muscle patches 
that actuate in antiphase (sine and cosine) at the same frequency and amplitude.
The actuation state of each particle,
$\omega_{\varphi, t}$,
was computed as the harmonic addition of the two channels of actuation (i.e.~sine and cosine waves).
The contribution from each channel of actuation was computed as described in Eq.~\ref{eq:actuation}.
\\

\refstepcounter{subsection}
\subsection*{4.6. Object manipulation (Fig.~4E-G).}
\label{sec:ObjectManipulation}

A round object 
of 208 particles
(radius 2.5 cm)
was added to the simulation, 
just above the top of the robot, and centered front to back. 
The robot was then optimized for:
locomotion 
(maximize robot velocity; Fig.~4E),
object transportation 
(maximize average velocity of robot and object; Fig.~4F);
and object ejection (maximize velocity of object, Fig.~4G).
\\

\refstepcounter{subsection}
\subsection*{4.7. Interactive design (Fig.~4H-L).}
\label{sec:MaskedDesigns}

The assumption of an initial rectangular grid was relaxed and replaced with any arbitrary user-supplied 2D shape, such as triangles (Fig.~4H,I), a circle (Fig.~4J), a heptagram (Fig.~4K) or enneagram (nine-sided star; Fig.~4L).
\\

\refstepcounter{subsection}
\subsection*{4.8. Constrained actuation (Fig.~\ref{fig:S4ConstrainedActuation}).}
\label{sec:ConstrainedActuation}

Manufacturing the automatically-designed robot 
required 
a post-processing step to be applied to the simulated design after optimization
(described above in Sect.~\ref{sec:physicalExperiments}). 
An additional experiment was thus conducted to determine if this post-processing step could instead be incorporated into the simulation 
\textit{ab initio} 
as a optimization constraint, 
so that simulated designs could directly be instantiated with closed-bladder actuators. 
To achieve this, particles close to the border were forbidden from actuating even when they were close to a patch of muscle (Fig.~\ref{fig:S4ConstrainedActuation}A). 
To resolve the passive border, 
the resolution of the simulated robot 
was increased by a factor of two
(128 particles in width and 88 in height; a total of 11264 particles). 
Starting with the same random initial configuration, 
a nearly identical morphology emerged after ten design attempts 
with (Fig.~\ref{fig:S4ConstrainedActuation}C) 
and without (Fig. 1C-L) 
constrained musculature. 
A slight yet significant increase in fitness was observed \textit{in silico} when actuation was constrained ($p < 0.01$); 
however, it is unclear whether or not this is an artifact of the higher particle count (11264 vs 2816 particles).
\\

\refstepcounter{subsection}
\subsection*{4.9. Direct particle optimization (Fig.~\ref{fig:S5DirectParticle}).}
\label{sec:DirectParticleOptimize}

The necessity of internal voids for optimizing robot shape can be seen when voids are replaced by a particle-based representation that directly stores and optimizes the mass and elasticity of each particle using gradient descent. 
Holding all other variables constant, including the use of muscle patches, the direct particle representation yielded robots with significantly lower fitness than that of robots optimized via explicit voids $(p < 0.01$; Fig.~\ref{fig:S5DirectParticle}A). 
This held true even when the direct particle representation was allowed 40 additional design evaluations ($p < 0.01$).
Moreover, without internal voids, shape change did not occur. 
In every one of the 25 independent trials of direct particle optimization, all initial particles remained part of the body after 50 design evaluations. 
This is significantly different ($p < 0.01$) to the void-based representation which typically reduces the total present particle count by 6\% over the first 10 design attempts (Fig.~\ref{fig:S5DirectParticle}B).
\\

\refstepcounter{subsection}
\subsection*{4.10. Intra-void interpolation (Figs.~\ref{fig:S6VoidInterpolation} and \ref{fig:S8RobustnessToHyperparameters}E).}
\label{sec:VoidInterpolation}

Each internal void contains 
a ``soft fringe'' 
that gently interpolates 
erasure, quadratically, 
from outer edge to mid-center
(Figs. 2C and \ref{fig:S7RobustnessToHyperParametersKey}), 
prior to fully removing the particles 
below a mass threshold 
at their inner core (Sect.~\ref{sec:ThresholdForRemoval}). 
Some kind of interpolation 
is required 
along 
a void's fringe
as 
a binary step function
(fully present particles encircling completely removed particles)
would not allow gradients to propagate 
from simulation 
to the design parameters
(void size and position). 
This is because, 
with a step function, 
given an $\epsilon >0$,
there does not exist any $\delta > 0$ where a change of the design parameters by an amount of $\delta$ always results in every particle changing its mass by an amount of less than or equal to $\epsilon$. 
This property is required for backpropagation.
\\

To evaluate the utility of the \textit{soft} fringe, 
additional experiments were conducted in which 
this 
quadratic interpolation 
was replaced with 
linear
and 
cubic interpolation regimes (Fig.~\ref{fig:S7RobustnessToHyperParametersKey}C). 
Under these regimes the mass of each particle is set to:
\begin{align}
    d =& \min_{k\in K} (d_{\varphi,k}^*, 1) \label{eq:void_interp_a}\\
    m_\varphi =& \begin{cases}
        0 &\text{if } d^q < \lambda \\
        d^q &\text{otherwise,}
    \end{cases}\label{eq:void_interp_b}
\end{align}
with $q \in \{1,2,3\}$ yielding linear, quadratic, and cubic interpolation, respectively (Fig.~\ref{fig:S8RobustnessToHyperparameters}E),
for particles within voids with mass above the threshold of particle inclusion $\lambda \in \{ 0.1, 0.2, 0.4\}$ (Fig.~\ref{fig:S8RobustnessToHyperparameters}G).
In Fig.~\ref{fig:S6VoidInterpolation}B, $\lambda$ was set to $0.5$ for the linear interpolation function.
\\

The fitness of designs with the soft (quadratic) fringe was significantly higher than that of the hard (linear) fringe ($p < 0.01$).
However there was not an appreciable difference in fitness
for designs optimized under
quadratic and cubic interpolation regimes.
While a soft fringe
facilitated 
optimization under the tested conditions,
it remains to determine precisely how
different interpolation regimes
alter 
the search landscape 
and gradients.
\\

\refstepcounter{subsection}
\subsection*{4.11. Intra-muscle interpolation (Fig.~\ref{fig:S8RobustnessToHyperparameters}F).}
\label{sec:MuscleInterpolations}

To evaluate the importance of the original quadratic interpolation within each actuator patch, 
additional experiments were conducted in which the quadratic interpolation was replaced with other interpolation regimes. 
Under these regimes actuation amplitude of each particle was set to:
\begin{equation}
    \label{eq:actuationInterpolation}
    A_{\varphi} = (1 - (d_{\varphi,c}^* \times \sqrt{0.1}))^q,
\end{equation}
where $q \in \{1,2,3\}$. 
Gradient descent does not appear to be sensitive to this 
(or any other tested) 
hyperparameter
and retains supremacy over previous approaches (solid versus dotted lines in Fig.~\ref{fig:S8RobustnessToHyperparameters}) in all regimes.
Note that
different muscle interpolations 
can generate 
very different actuation signatures across the robot's body.
So, we should be careful about drawing conclusions about which interpolation regime is the best.
\\

\refstepcounter{subsection}
\subsection*{4.12. Threshold for removal (Fig.~\ref{fig:S8RobustnessToHyperparameters}G)} 
\label{sec:ThresholdForRemoval}

Particles 
along the outer fringe of voids 
decrease in physicality (mass and elasticity; Sect.~\ref{sec:VoidInterpolation})
until their mass falls below a threshold,
$\lambda$,
after which 
they are completely removed from simulation.
The physical robot was 
optimized in simulation 
with $\lambda=0.1$ kg (Fig.~2).
Later, two higher ($\lambda=0.2,0.4$ kg) thresholds were explored (Fig.~\ref{fig:S8RobustnessToHyperparameters}G).
Fitness slightly (but significantly) decreased when the threshold was raised to 0.2 kg
and slightly (but significantly) decreased further when threshold was raised to 0.4 kg.
However,
different thresholds 
have 
different 
numbers of particles 
and different
material properties.
So, this comparison is not ``apples to apples'' and does not suggest smaller thresholds are better.
It does show, however, that gradient descent is not
overly sensitive to changes in this hyperparameter, 
which is interesting since it 
directly affects the gradients.
\\


\refstepcounter{subsection}
\subsection*{4.13. Initial void radius distribution (Figs.~\ref{fig:S8RobustnessToHyperparameters}A)}
\label{sec:VoidRadiiDistributions}

The distribution of void radii used to design the physical robot 
(Sect.~\ref{sec:optimization})
was replaced
by
one of eight other distributions (Fig.~\ref{fig:S7RobustnessToHyperParametersKey}A):
$\text{Unif}(0,2a),\; 
\text{Unif}(a-a^2,a+a^2),\; 
\text{Unif}(0,2b),\; 
\text{Unif}(b-b^2,b+b^2),\;
\text{Norm}(a,a),\; 
\text{Norm}(a,a^2),\; 
\text{Norm}(b,b),\; 
\text{Norm}(b,b^2)$.
When $|K|$ voids are sampled from
one of these distributions and 
arranged side-by-side on a grid,
their \textit{grid coverage},
\begin{equation}
O_K = \frac{\pi}{W} \sum_{k \in K} r_k^2 \;,  
\label{eq:gridCoverage}
\end{equation}  
will be either
$O_K=30\%$ (when $\mu=a$) 
or $O_K=60\%$ (when $\mu = b$)
of the robot's body area, $W$, on average.
\\


\refstepcounter{subsection}
\subsection*{4.14. Initial muscle radius distribution (Fig.~\ref{fig:S8RobustnessToHyperparameters}B)} 
\label{sec:MuscleRadiiDistributions}

The muscle radius used to design the physical robot
(Sect.~\ref{sec:optimization})
was replaced by 
radii 
sampled 
from
one of the same eight distributions used for voids in Sect.~\ref{sec:VoidRadiiDistributions}.
That is:
$\text{Unif}(0,2a),\; 
\text{Unif}(a-a^2,a+a^2),\; 
\text{Unif}(0,2b),\; 
\text{Unif}(b-b^2,b+b^2),\;
\text{Norm}(a,a),\; 
\text{Norm}(a,a^2),\; 
\text{Norm}(b,b),\; 
\text{Norm}(b,b^2)$.
When $|C|$ muscles are sampled from one of these
distributions and 
arranged side-by-side on a grid,
their 
grid coverage, $O_C$, 
which sums muscle areas 
in a similar fashion as Eq.~\ref{eq:gridCoverage},  
will be either
$O_C=57.5\%$ (when $\mu=a$) 
or $O_C=115\%$ (when $\mu = b$)
of the robot's body area, on average.
In practice, however, 
the muscles always overlap.
\\


\refstepcounter{subsection}
\subsection*{4.15. Number of voids (Fig.~\ref{fig:S8RobustnessToHyperparameters}C)}
\label{sec:NumberOfVoids}

The $|K|=64$ voids used to design the physical robot were 
replaced by
8,
16,
32,
128,
and 256
voids 
(Fig.~\ref{fig:S7RobustnessToHyperParametersKey}B).
\\


\refstepcounter{subsection}
\subsection*{4.16. Number of muscles (Fig.~\ref{fig:S8RobustnessToHyperparameters}D)} 
\label{sec:NumberOfMuscles}

The $|C|=64$ muscles used to design the physical robot were 
replaced by
8,
16,
32,
128,
and 256
muscles (Fig.~\ref{fig:S7RobustnessToHyperParametersKey}B).
\\ \\


\addcontentsline{toc}{section}{\protect\numberline{}SI References.}%


\section*{Figures:}
\addcontentsline{toc}{section}{\protect\numberline{}Figures S1 to S13.}%

\begin{figure}
    \centering
    \includegraphics[width=\linewidth]{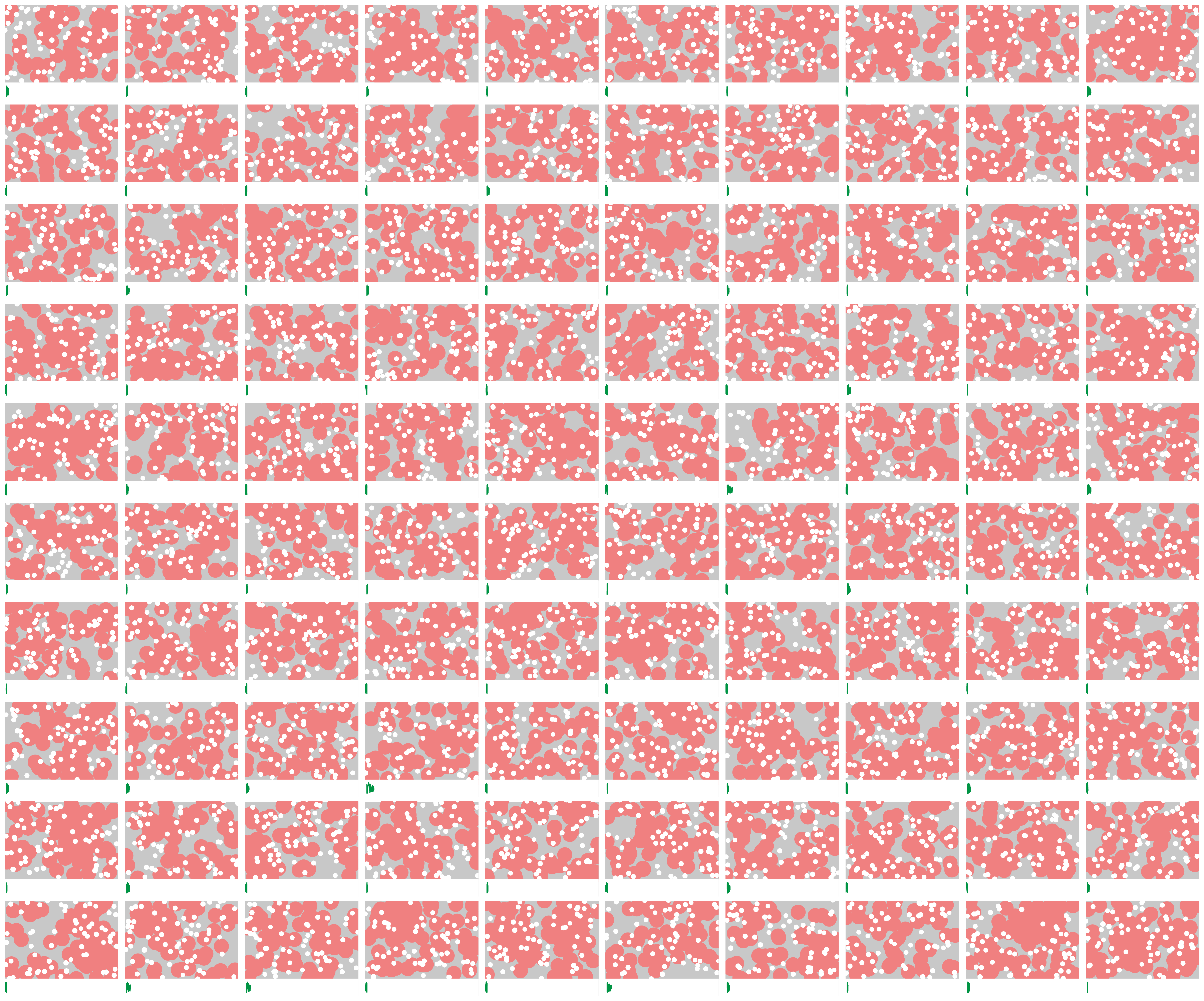}
    \caption{\textbf{Attempt 1}. All 100 random initial configurations (first attempts) across 100 independent trials. Green CoM traces indicate that the initial designs primarily bounce up and down in place.}
    \label{fig:S2Initial100Designs}
\end{figure}

\begin{figure}
    \centering
    \includegraphics[width=\linewidth]{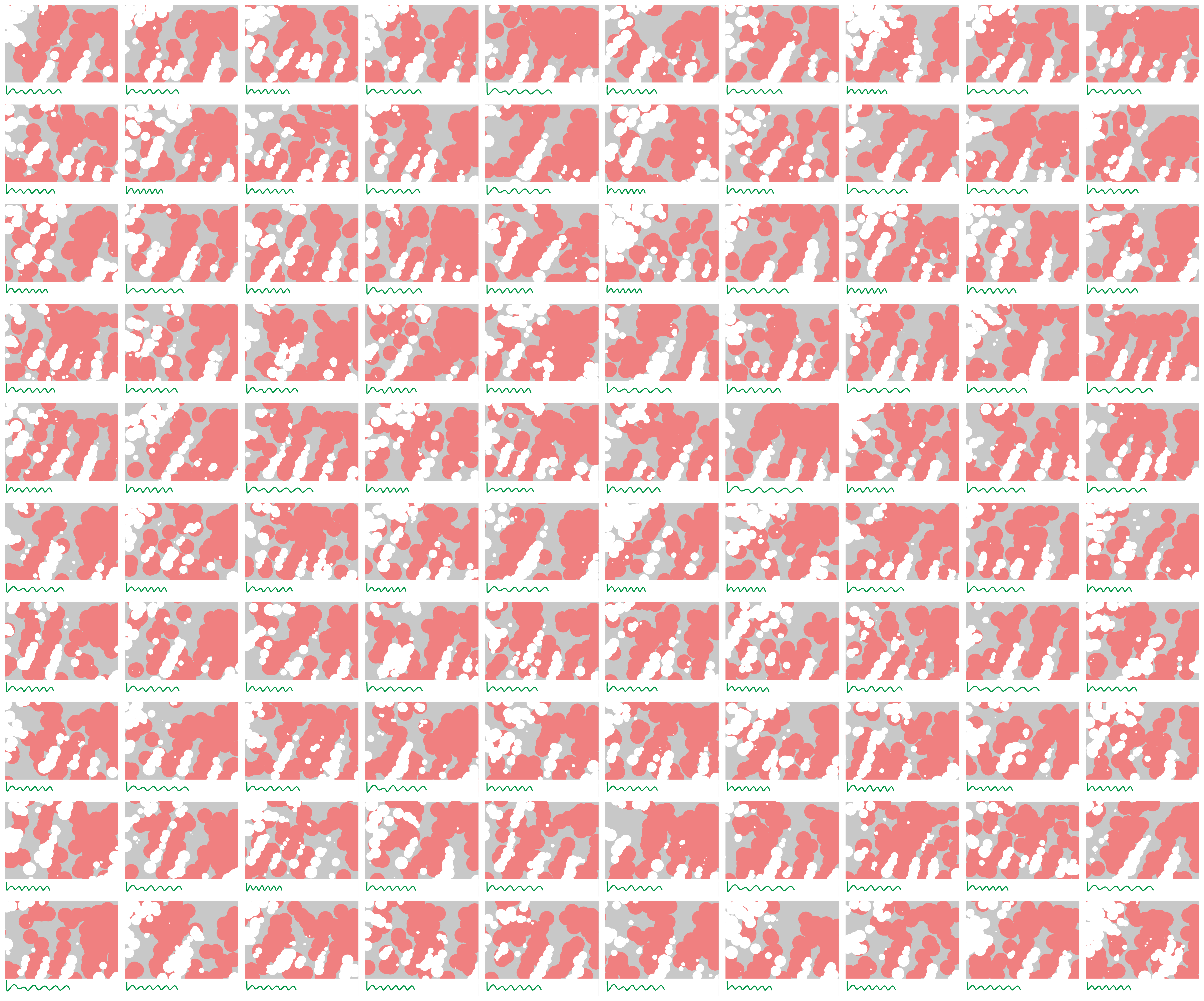}
    \caption{\textbf{Attempt 10}. Optimized designs from the 100 independent trials starting with the 100 different random initial configurations (random seeds) in Fig.~\ref{fig:S2Initial100Designs}. Green CoM traces indicate all optimized designs were able to locomote forward with stable gaits. Morphological convergence to posterior-angled legs with anterior musculature can be observed across these 100 independent trials. However, this design principle is embodied in the 100 designs in 100 different ways, each with their own unique body shape, number of limbs (2 to 4), and overall musculature.}
    \label{fig:S3Optimized100Designs}
\end{figure}

\begin{figure}
    \centering
    \includegraphics[width=\linewidth]{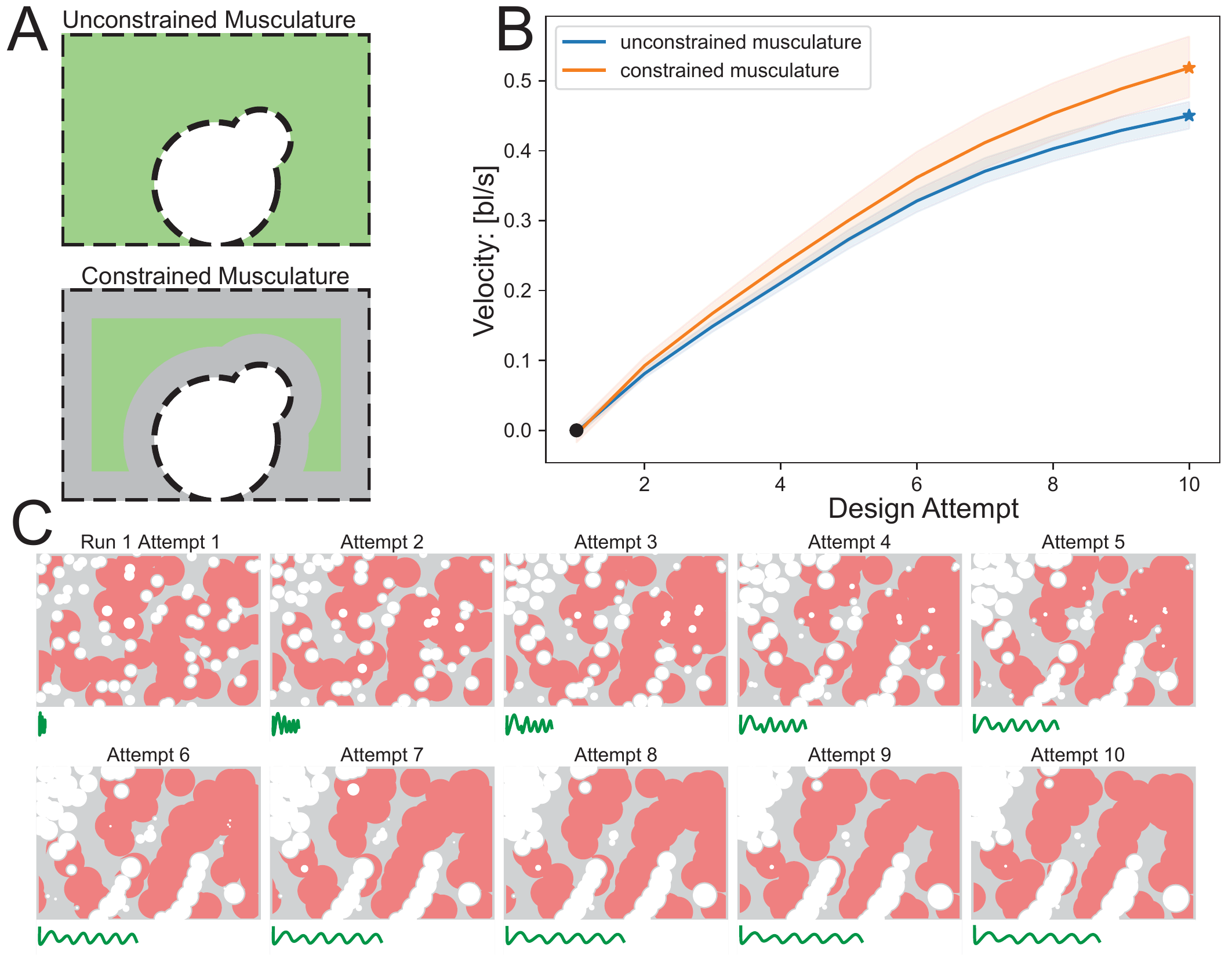}
    \caption{\textbf{Design with constraints}. Morphology was constrained such that muscles are buildable as closed pneumatic bladders (\textbf{A}). This did not render the body less differentiable in preliminary experiments in terms of locomotion ability. In fact there seems to be a slight but significant improvement in fitness (\textbf{B}). However, this may be an artifact stemming from the use of four times as many particles used to simulate constrained actuation (higher resolution is needed to resolve the passive border; Table.~\ref{table:design}). (\textbf{C}:) Ten design attempts starting from the same initial design parameters as in Fig. 1C. The final constrained actuation design looks visually similar to the final unconstrained design from Figure 1L.
}
    \label{fig:S4ConstrainedActuation}
\end{figure}

\begin{figure}
    \centering
    \includegraphics[width=\linewidth]{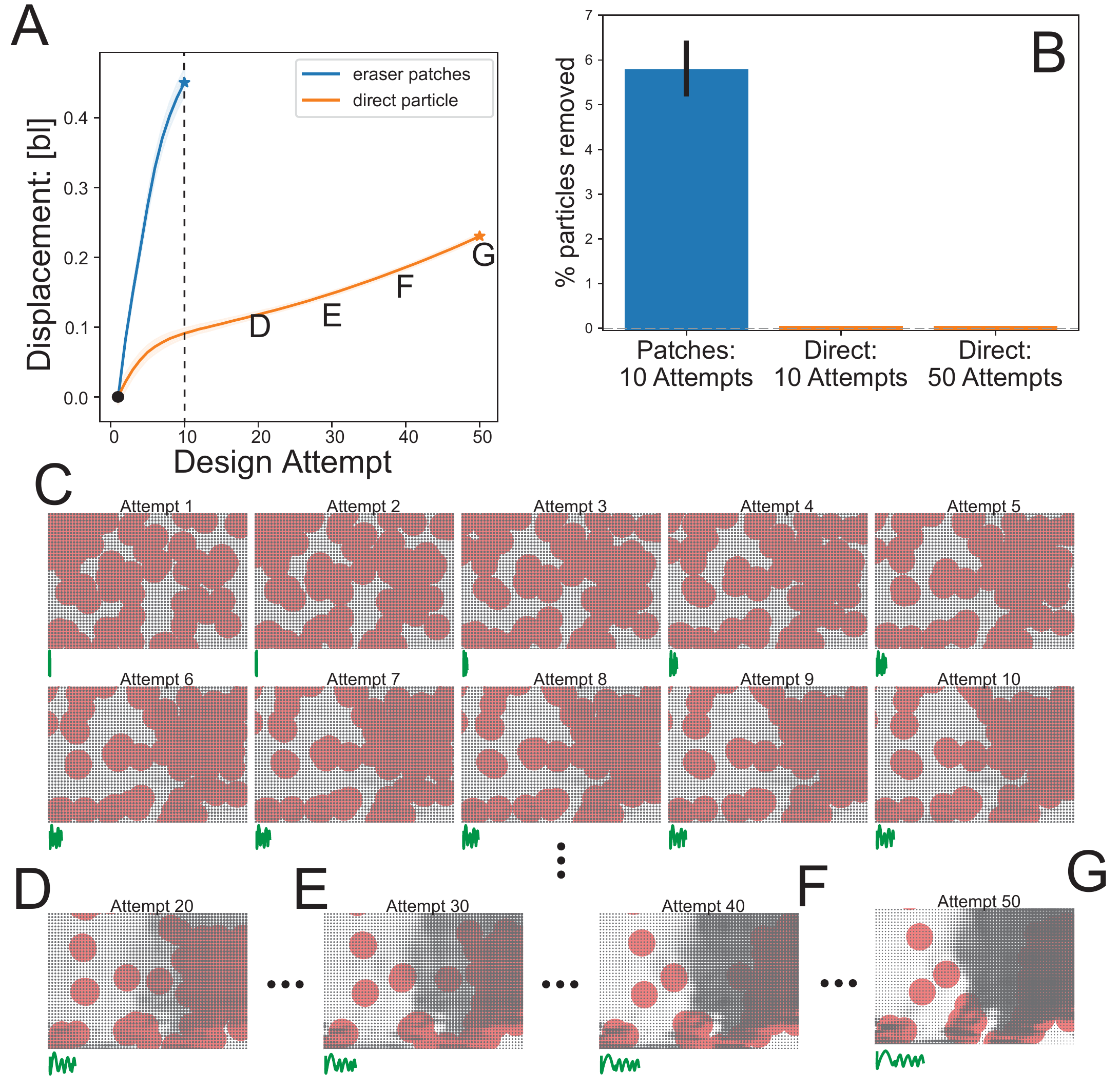}
    \caption{\textbf{The necessity of explicit voids}. Without internal voids, fitness was greatly diminished ($p < 0.01$; \textbf{A}), and shape change did not occur (\textbf{B}). Design attempts 1 through 10 exhibit show minimal changes in both material properties and fitness (\textbf{C}). Further design attempts (attempts 20 to 50; \textbf{D-G}) hold on to the same rectangular body plan and actuation placements, and they exhibit unstable gaits.}
    \label{fig:S5DirectParticle}
\end{figure}

\begin{figure}
    \centering
    \includegraphics[width=\linewidth]{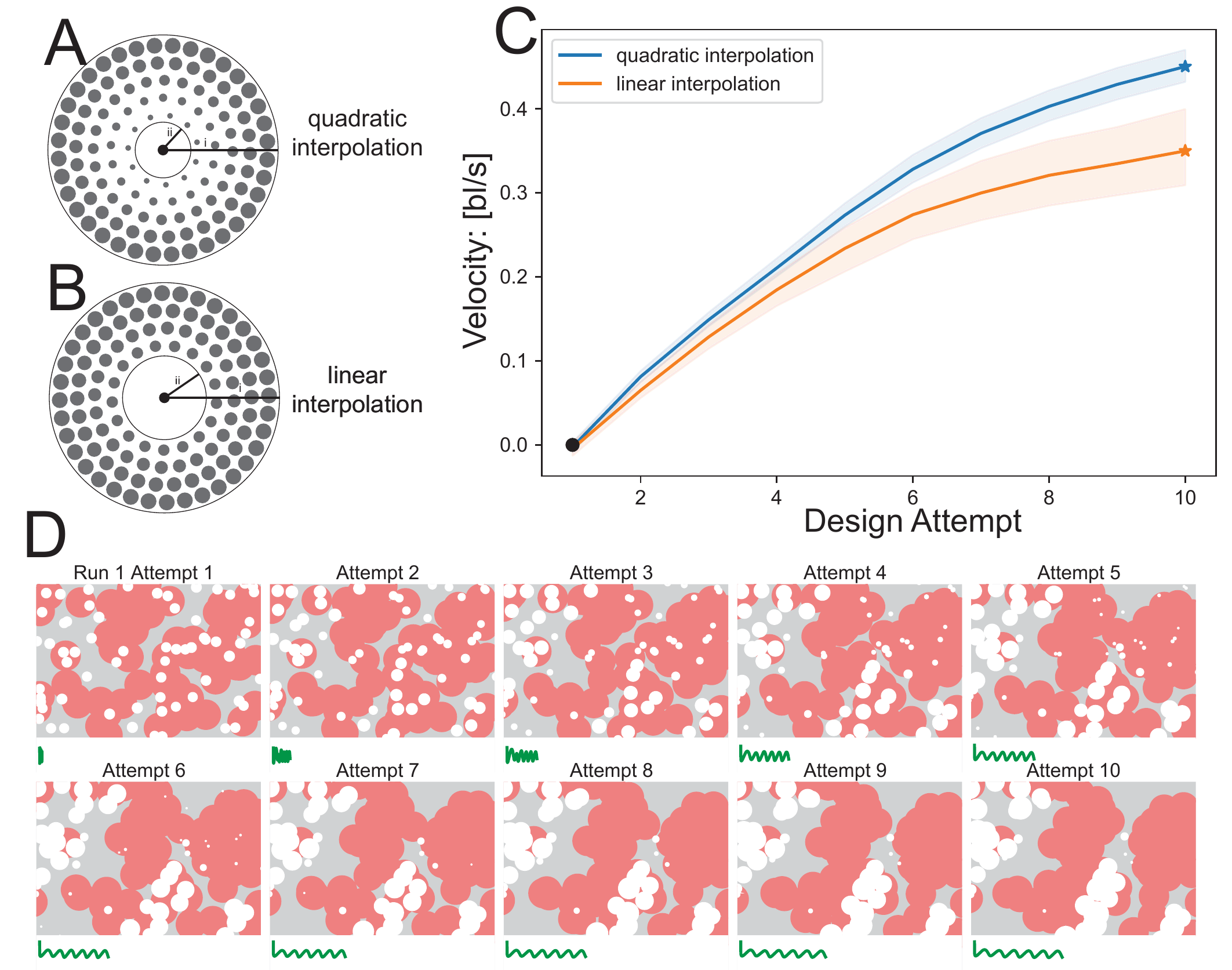}
    \caption{\textbf{The utility of intra-void interpolation}. 
    A smooth (quadratic) interpolation was used between the edge and center of each void ($q=2, \lambda=0.1$; \textbf{A}), 
    which
    rendered the body more differentiable than a sharper (linear) interpolation ($q=1,\lambda=0.5$; \textbf{B}) in terms of the fitness achieved after ten design attempts (\textbf{C}). 
    Morphologies optimized with more sharply interpolated voids (\textbf{D}) were very similar to those with smoother interpolation: 
    both have posterior-angled legs with anterior musculature.}
    \label{fig:S6VoidInterpolation}
\end{figure}

\begin{figure}
    \centering
    \includegraphics[width=\linewidth]{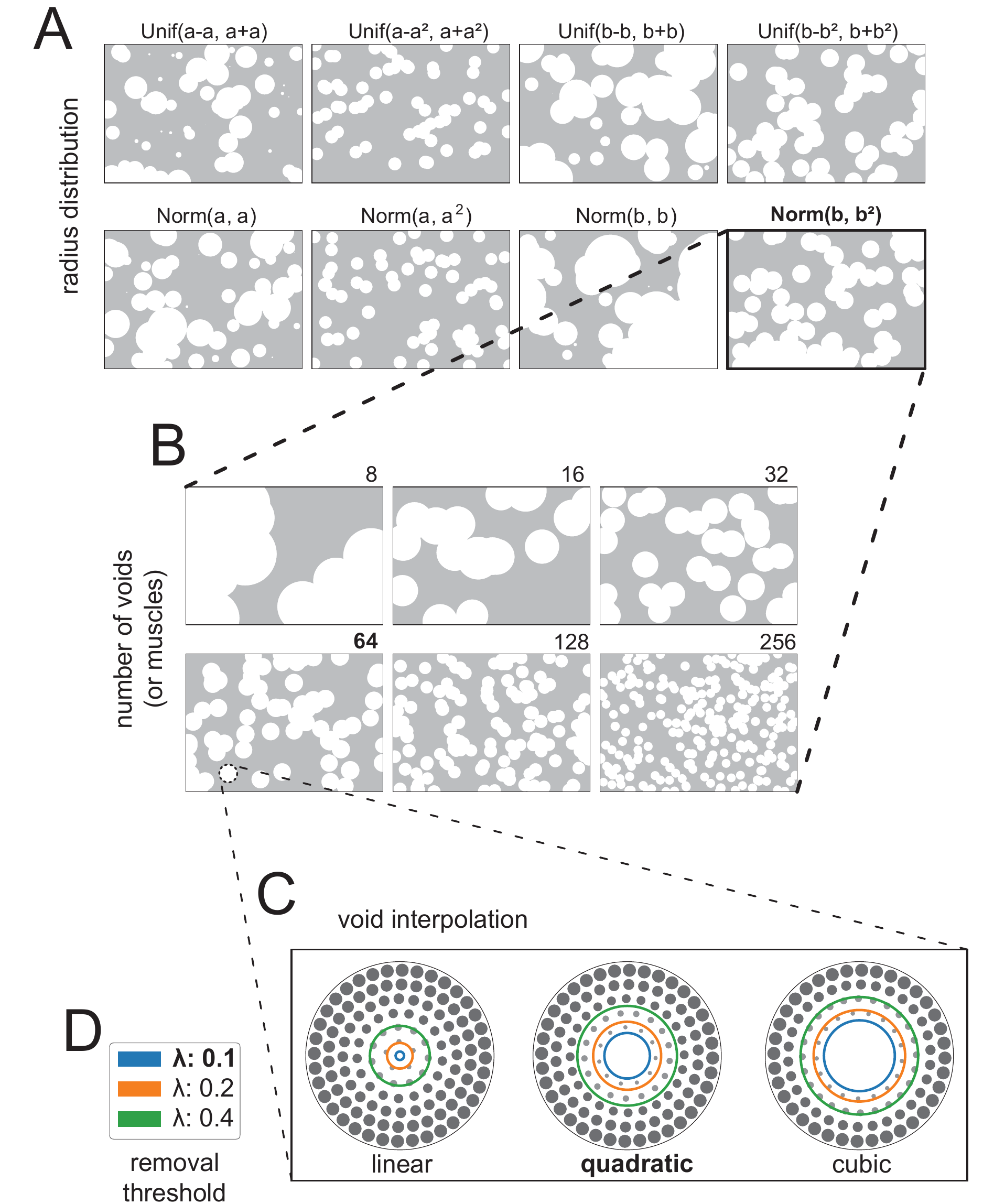}
    \caption{
    \textbf{Summary of hyperparameter sweep.}
    We repeated our experiment under tens of thousands of different conditions: 
    different combinations of hyperparameter values 
    (detailed in Table~\ref{table:design}).
    These hyperparameters 
    determine
    the distribution of initial radius 
    for voids and muscles (\textbf{A}),
    the number of voids and muscles (\textbf{B}), 
    the interpolation within voids (\textbf{C}) and muscles,
    and the
    threshold for particle removal within voids (\textbf{D}).
    The original hyperparameters used to optimize
    the physical robot (Figs.~1-3) are bolded.
    The
    results of the hyperparameter sweep
    can be seen in Figs.~\ref{fig:S8RobustnessToHyperparameters} and \ref{fig:S9RobustnessToHyperparametersGallery}.
    }
    \label{fig:S7RobustnessToHyperParametersKey}
\end{figure}

\begin{figure}
    \centering
    \includegraphics[width=0.9\linewidth]{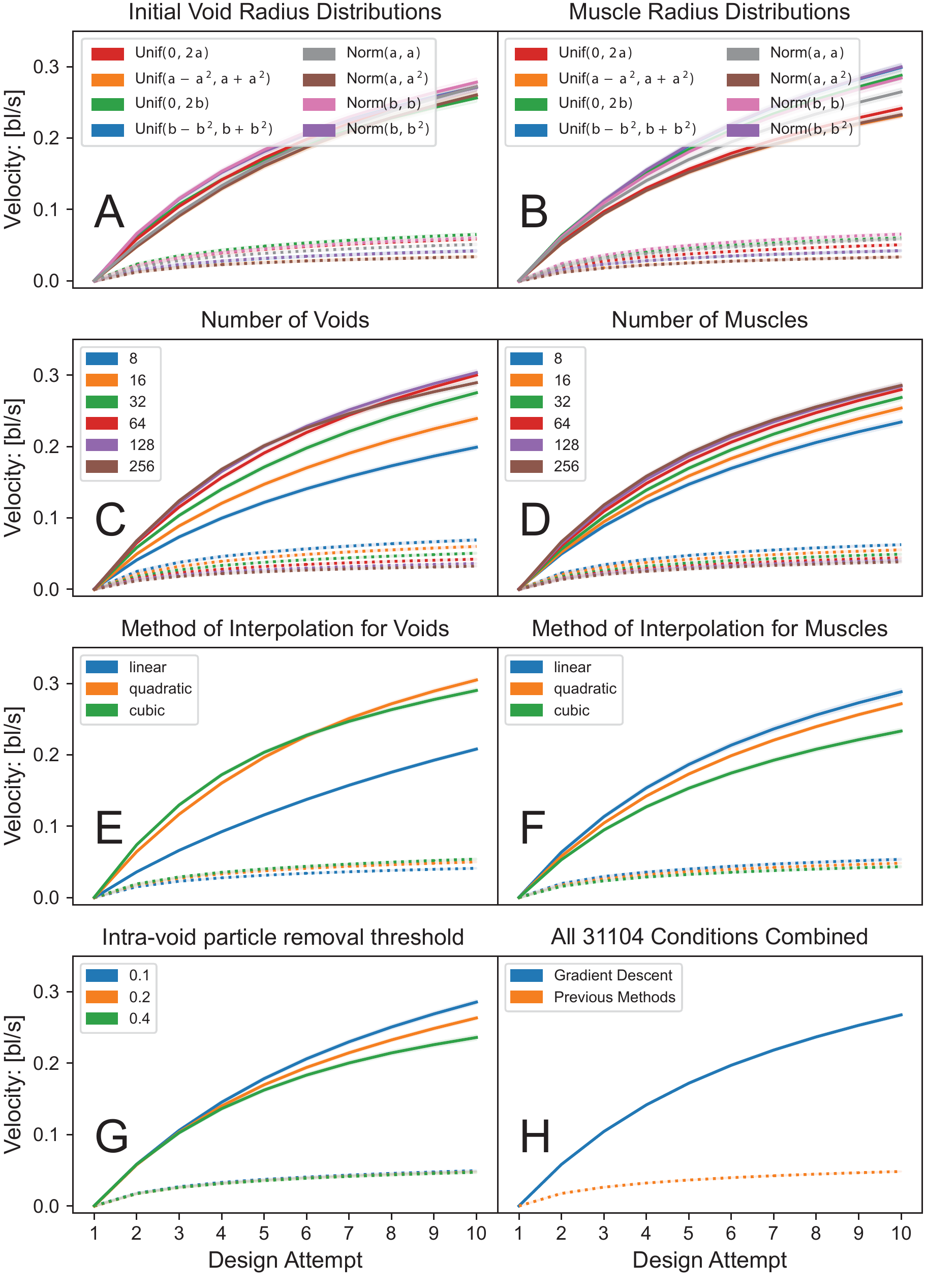}
    \caption{\textbf{Robustness to hyperparameters}. We repeated our experiment under 31104 different conditions, spanning eight different distributions of initial radius (four uniform and four normal distributions, 
    with different locations and scales,
    whose means
    correspond to
    $|K|$ sampled voids 
    with areas
    that sum to 30\% of the robot’s body area, when $\mu=a$, 
    or 60\%, when $\mu=b$; \textbf{A,B}), 
    six different numbers of voids (8, 16, 32, 64, 128, 256; \textbf{C}), 
    six different numbers of muscles (8, 16, 32, 64, 128, 256; \textbf{D}), 
    three different methods of void interpolation (linear, quadratic, cubic; \textbf{E}), 
    three options of muscle interpolation (linear, quadratic, cubic; \textbf{F}), 
    and three different thresholds for removal of material (10\%, 20\%, 40\%; \textbf{G}), 
    comparing gradient descent (solid lines) to previous methods (dotted lines) with a budget of 10 design attempts (\textbf{H}).
    Means are plotted with 99\% normal confidence intervals with Bonferroni correction for 38 comparisons (solid lines vs dotted lines).}
    \label{fig:S8RobustnessToHyperparameters}
\end{figure}

\begin{figure}
    \centering
    \includegraphics[width=\linewidth]{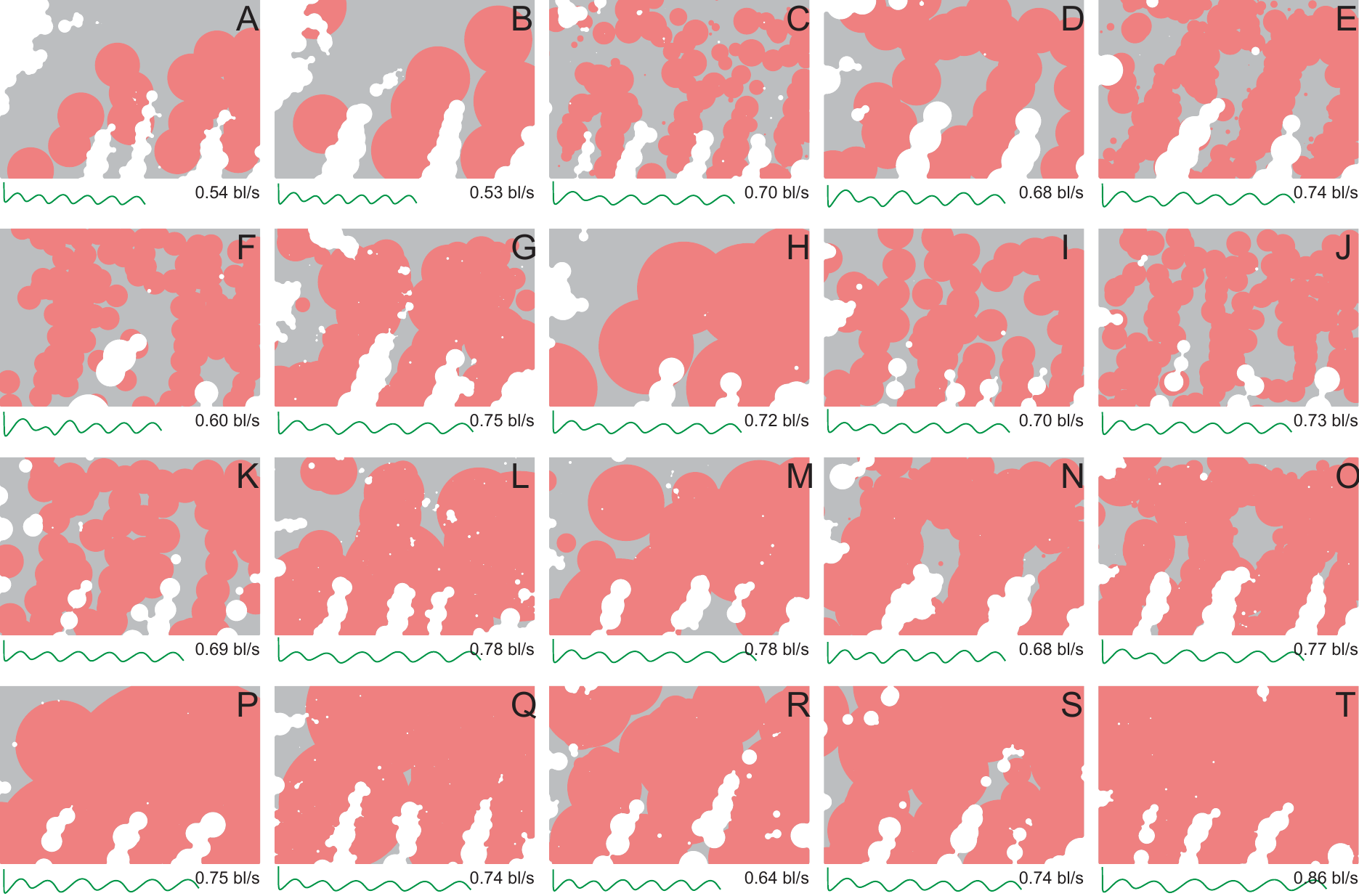}
    \caption{\textbf{Locally optimal robots}. (\textbf{A-T}:) The most fit robot found by gradient descent in each of the 38 comparisons of Fig.~\ref{fig:S8RobustnessToHyperparameters} Because these comparisons aggregate results across overlapping sets of hyperparameter values, the best overall robot (T) was the best in 8 of the 38 comparisons. Other high performing robots also dominated multiple comparisons, yielding the 20 unique individuals plotted here. Gradient descent converged with high probability to local peaks in the non-convex search landscape: Distinct morphologies—different body shapes, numbers of limbs (3 to 5), and musculature (red)—capable of legged locomotion (green curves trace the robot’s CoM during behavior). Designs are sorted from least (A) to most (T) muscle coverage.
}
    \label{fig:S9RobustnessToHyperparametersGallery}
\end{figure}

\begin{figure}
    \centering
    \includegraphics[width=\linewidth]{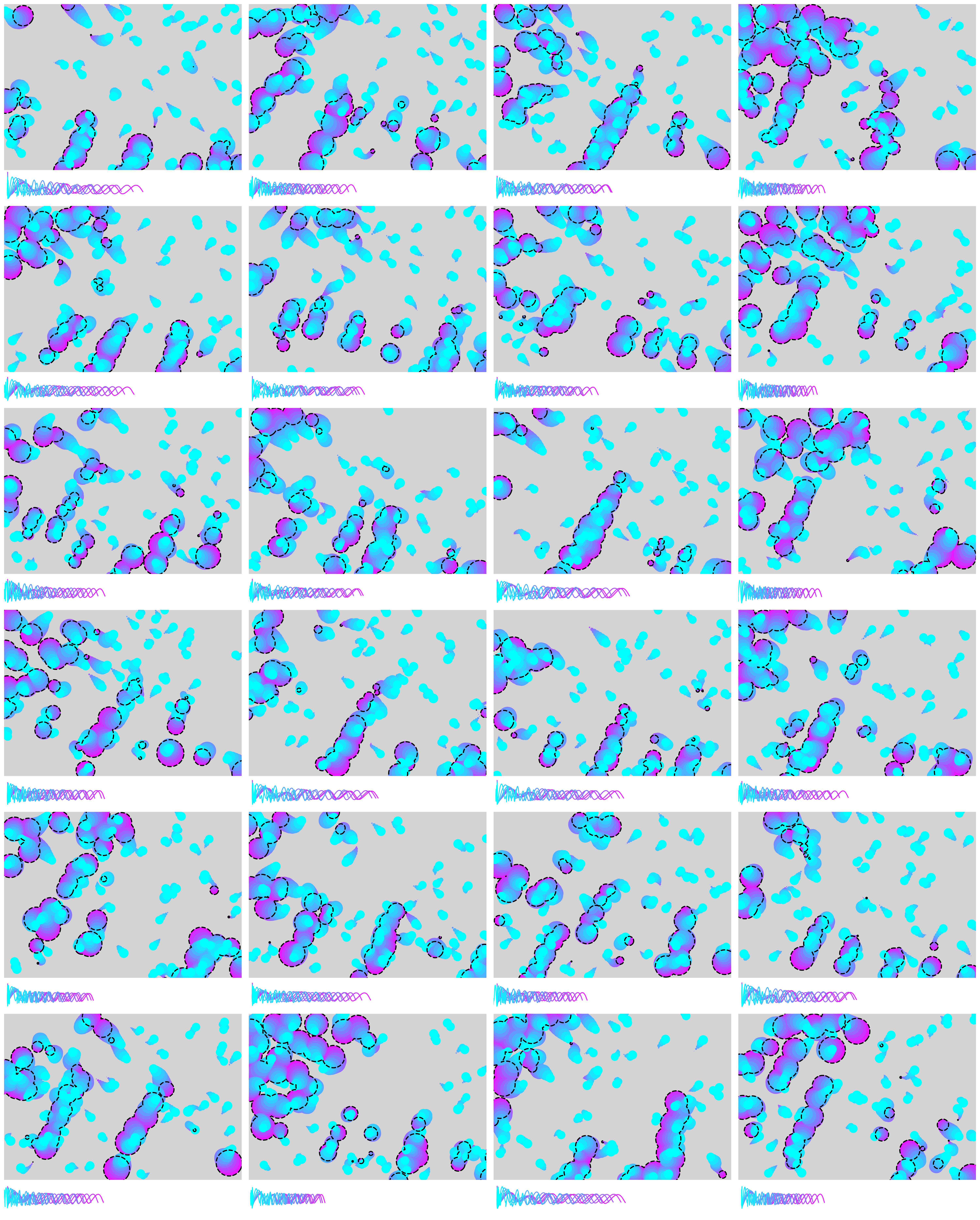}
    \caption{\textbf{Shape change (runs 5-28)}. (Runs 1-4 can be found in Fig. 3.) Position and size of the internal voids from the random initial configurations (cyan circles) to final optimized body plan (purple circles; outlined by a black dashed line) for independent trials 5 through 28. For each run, the behavior is shown for each of the 10 design attempts (cyan to purple waves).}
    \label{fig:S10designdevelopment528}
\end{figure}

\begin{figure}
    \centering
    \includegraphics[width=\linewidth]{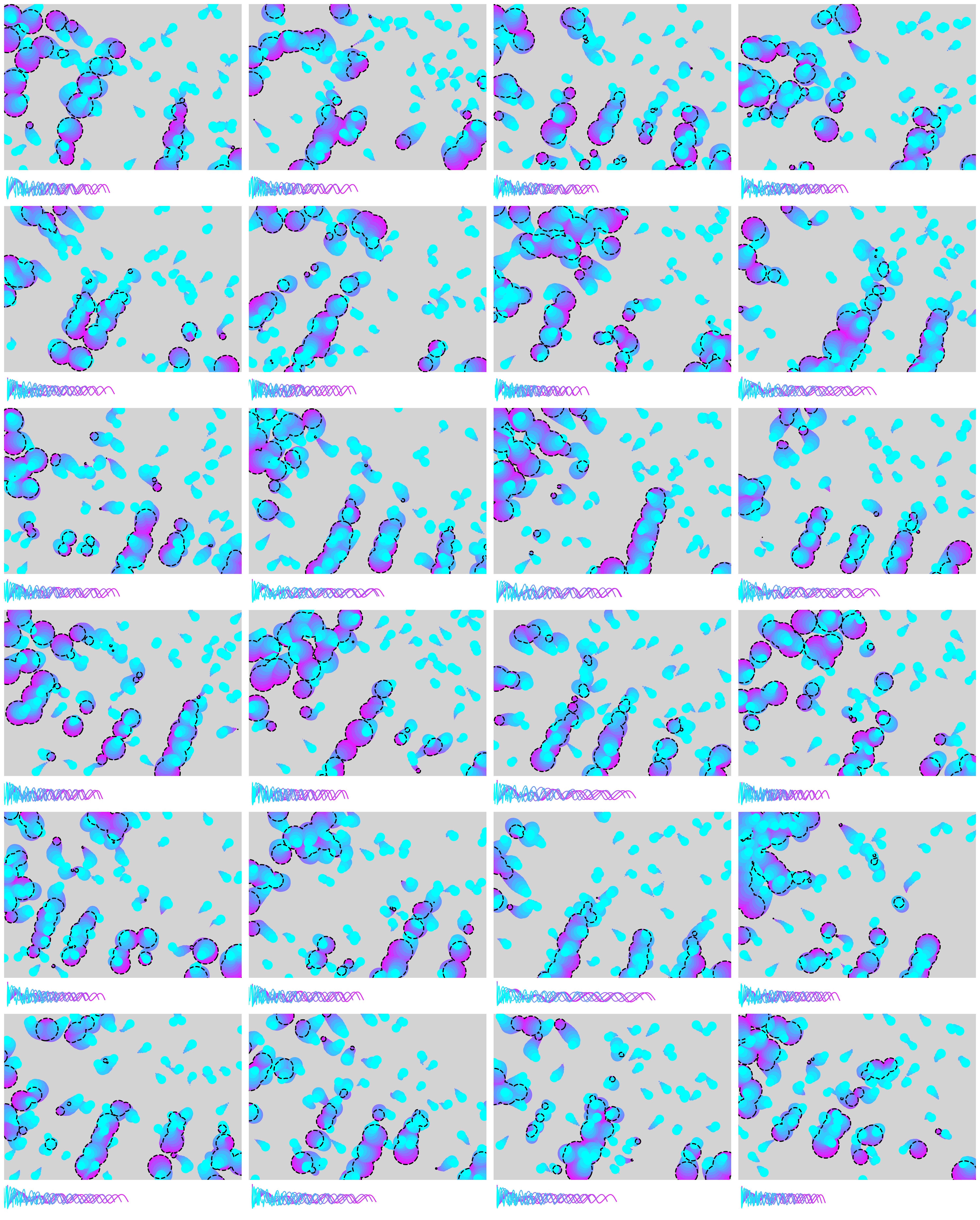}
    \caption{\textbf{Shape change (runs 29-52)}. Position and size of the internal voids from the random initial configurations (cyan circles) to final optimized body plan (purple circles; outlined by a black dashed line) for independent trials 29 through 52. For each run, the behavior is shown for each of the 10 design attempts (cyan to purple waves).
}
    \label{fig:S11designdevelopment2952}
\end{figure}

\begin{figure}
    \centering
    \includegraphics[width=\linewidth]{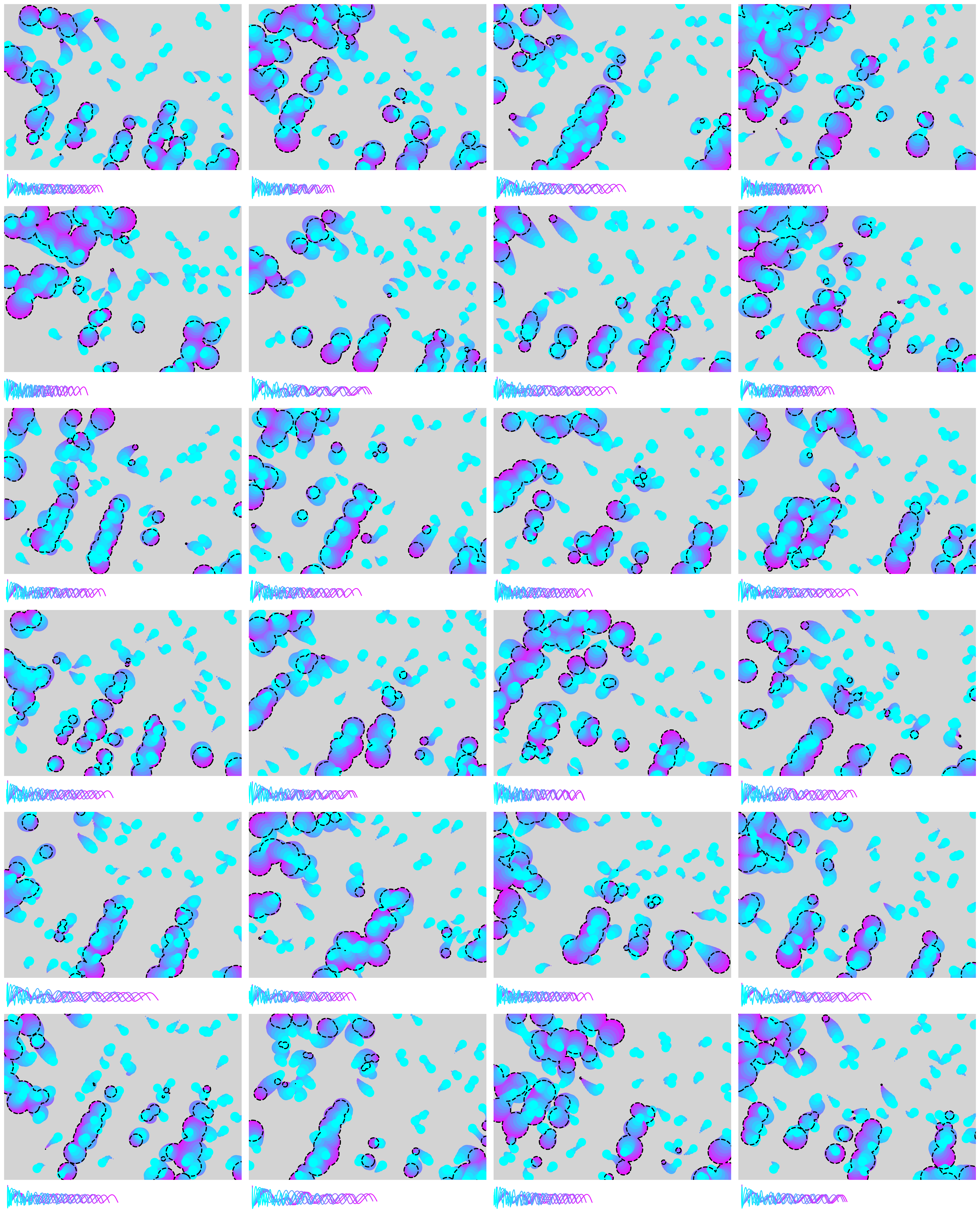}
    \caption{\textbf{Shape change (runs 53-76)}. Position and size of the internal voids from the random initial configurations (cyan circles) to final optimized body plan (purple circles; outlined by a black dashed line) for independent trials 53 through 76. For each run, the behavior is shown for each of the 10 design attempts (cyan to purple waves).}
    \label{fig:S12designdevelopment5376}
\end{figure}

\begin{figure}
    \centering
    \includegraphics[width=\linewidth]{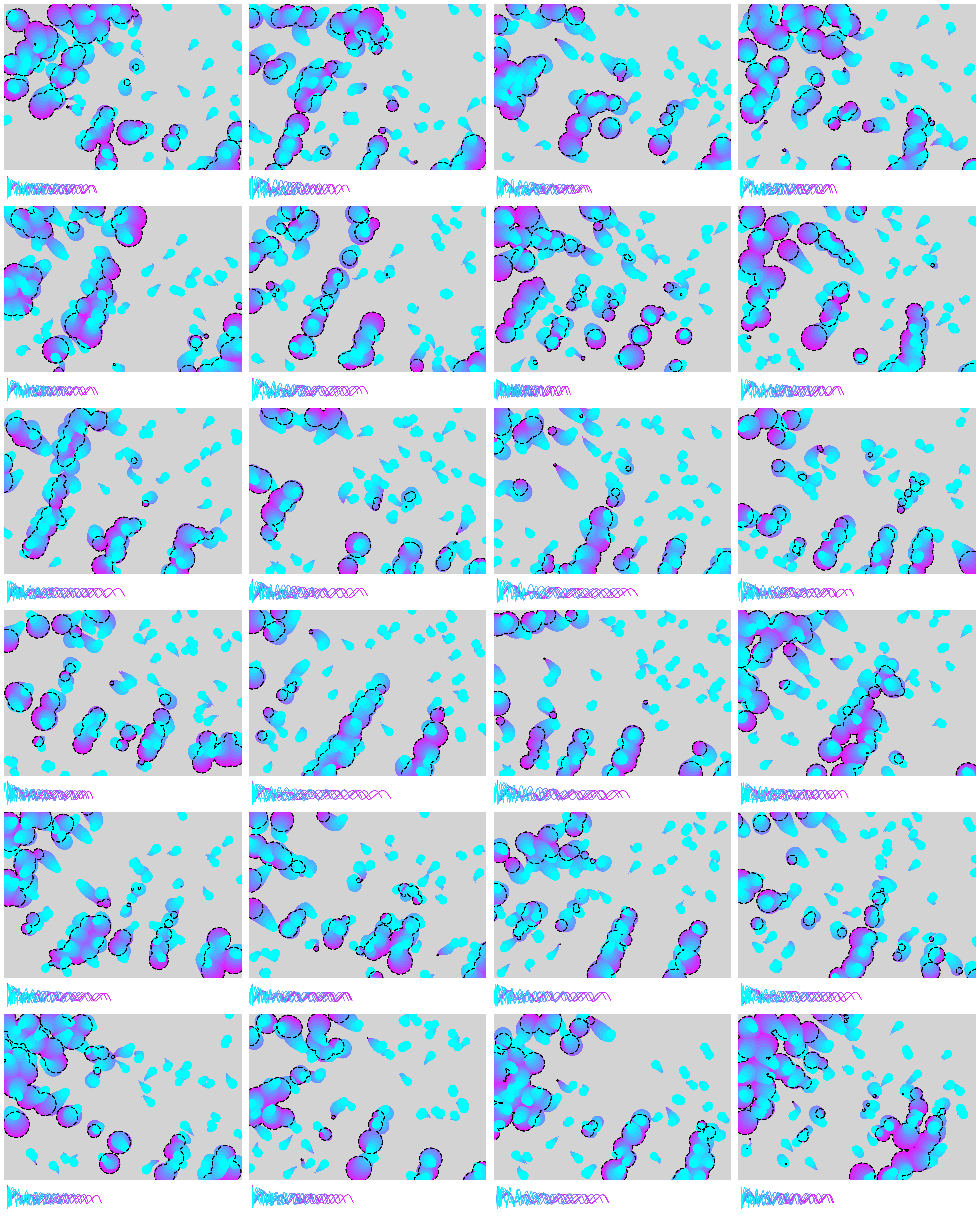}
    \caption{\textbf{Shape change (runs 77-100)}. Position and size of the internal voids from the random initial configurations (cyan circles) to final optimized body plan (purple circles; outlined by a black dashed line) for independent trials 77 through 100. For each run, the behavior is shown for each of the 10 design attempts (cyan to purple waves).}
    \label{fig:S13designdevelopment77100}
\end{figure}

\section*{Tables:}
\addcontentsline{toc}{section}{\protect\numberline{}Tables S1 and S3.}%

\begin{table}[!h]
\caption{\textbf{Symbols.}
The following symbols are used above in Sects.~S1-S4.}
\begingroup
\renewcommand{\arraystretch}{1.33} 
\begin{tabular}{|c|l|}
\hline
    \textbf{Symbol} & \textbf{Description} \\ \hline
    $t\in\{1,2,\ldots,T\}$ & Timestep \\ \hline
    $dt$ & Step size  \\ \hline
    $W$ & Workspace length$\,\times\,$height [cm$^2$] \\ \hline
    $k \in K$ & Void  \\ \hline
    $c \in C$ & Muscle \\ \hline
    $r_k$ & Void radius \\ \hline
    $r_c$ & Muscle radius \\ \hline
    $O_K$ & grid coverage of $K$: summed void areas relative to robot's area.\\ \hline
    $O_C$ & grid coverage of $C$: summed muscle areas relative to robot's area. \\ \hline
    $\varphi \in \Phi$ & Particle  \\\hline
    $m_{\varphi}$ & Particle mass [kg]  \\\hline
    $\bar{m}_{\varphi}$ & Avg. particle mass (includes removed particles with zero mass) [kg]  \\\hline
    $E_{\varphi}$ & Particle elastic modulus [Pa] \\\hline
    $\Phi_{x,y}$ & Robot's center of mass (CoM)  \\ \hline
    $\lambda$ & Intra-void particle removal threshold [kg] \\ \hline
    $m^{\star}$ & Target average mass \\ \hline
    $A_{\varphi}$ & Particle actuation amplitude  \\\hline
    $\omega_{\varphi,t}$ & Actuation state of particle $\varphi$ at time $t$  \\\hline
    $f$ & Actuation frequency [Hz] \\ \hline
    $q$ & Intra-void/muscle interpolation power \\ \hline
    $D(\mathbf{x}_1,\mathbf{x}_2)$ & Euclidean dist. between cartesian coordinates $\mathbf{x}_1$ and $\mathbf{x}_2$ [cm] \\ \hline
    $d_{\varphi,c} = D(\varphi,c)$ & Euclidean dist. from $\varphi$ to $c$ [cm] \\ \hline
    $d_{\varphi,k} = D(\varphi,k)$ & Euclidean dist. from $\varphi$ to $o$ [cm] \\ \hline
    $d_{\varphi,c}^*$ or $d_{\varphi,k}^*$ & Normalized distance (dist.~divided by $r_c$ or $r_k$) \\ \hline
    $\nu_{\varphi} = D(\varphi,\Phi_{x,y})$ & Dist. of a particle to the robot's CoM 
    \\ \hline
    $\ell$ & Loss (negative fitness)  \\ \hline
    $\ell_{\text{sim}}$ & Simulator loss: the gradients from the differentiable physics simulation \\ \hline
    $\ell_{\text{addtl}}$ & Additional losses that were combined with $\ell_{\text{sim}}$ during optimization. \\ \hline
\end{tabular}
\centering
\endgroup
\label{table:symbols}
\end{table}

\begin{table}[t]
\caption{\textbf{Hyperparameters.} The following hyperparameters were used in the experiments detailed above in Sects.~S1-S4; 
those used to design the physical robot (Sect.~S4) are bolded.}
\begingroup
\renewcommand{\arraystretch}{1.8} 
\begin{tabular}{|l|l|}
\hline
\textbf{Variable}                                  & \textbf{Value}                                    \\ \hline
Particle count                                     & $\mathbf{64 \times 44  = 2816}$; 
                                                    and  $128 \times 88 = 11264$ (Fig.~\ref{fig:S4ConstrainedActuation}).
                                                    \\ \hline

Workspace length by height, $W$                    & $W = 20\times 14 = 280 \text{ cm}^2$                   \\ \hline
Void grid coverage, $O_K$                             & $O_K = \frac{\pi}{W} 
                                                        \sum r_k^2 
                                                        = 30\% \text{ or } \mathbf{60}\textbf{\%}$ \\ \hline
Muscle grid coverage, $O_C$                             &
                                                            $O_C = \frac{\pi}{W} 
                                                            \sum r_c^2 
                                                            \approx 57.5\% \text{ or } \mathbf{115}\textbf{\%}$                                              \\ \hline
Young’s modulus                            & 20 Pa                                             \\ \hline
Poisson’s Ratio                              & 0.25                                              \\ \hline
Actuation strength                                 & 4.0 Pa / particle                                 \\ \hline
Time steps ($T$)                                    & $T=1024$                                             \\ \hline
Step size ($dt$)                                     & $dt=0.001$ seconds                                     \\ \hline
MPM grid size                                      & $128\times128$                                           \\ \hline
Gravity                                            & $5.4 \text{m}/\text{s}^2$                        \\ \hline
Friction                                           & 0.5                                               \\ \hline
Internal viscous damping                           & 30.0                                              \\ \hline
Global viscous damping                             & 2.0                                               \\ \hline
Intra-void/muscle interpolation, $q$                      & $q \in \{1, \textbf{2}, 3\}$ \\ \hline
Threshold for removal, $\lambda$                      & $\lambda \in \{\textbf{0.1}, 0.2, 0.4\}\text{ kg}$                                     \\ \hline
Number of voids, $|K|$                             & $|K| \in \{8,16,32,\textbf{64},128,256\}$                  \\ \hline
Number of muscles, $|C|$                           & $|C| \in \{8,16,32,\textbf{64},128,256\}$                  \\ \hline
\makecell[l]{Initial mean void radius\\ \quad in silico (includes fringe), $r_k$}
                                                    & $r_k = \sqrt{\frac{O_K \times W }{\pi \times |K|}}\approx \mathbf{0.91}\;\textbf{cm}$ \\ \hline
Physical void radius (no fringe)                   & $r_k\sqrt{0.1}\times 1.3\approx                \mathbf{0.37}\;\textbf{cm}$ \\ \hline

Muscle radius, $r_c$                                & $r_c = \sqrt{\frac{O_C \times W }{\pi                                                             \times |C|}}\approx \mathbf{1.26}\;\textbf{cm}$
                                                      \\ \hline
                                                        
Actuation frequency, $f$                           & $f = \frac{40}{2\pi}\approx 6 \text{ Hz}$                                        \\ \hline
\makecell[l]{Initial void radius dist.} & \makecell[l]{
            $\{ \text{Norm}(a,a), \text{Norm}(a, a^2), \text{Norm}(b,b), \text{\textbf{Norm}}(\mathbf{b, b^2})$,  \\
            $\quad \text{Unif}(a-a, a+a), \text{Unif}(a - a^2, a+a^2), \text{Unif}(b-b, b+b),$  \\
            $\quad \text{Unif}(b - b^2, b+b^2)\}$,\\
            with 
            $a$ and $b$ corresponding to 
            $O_K=30\%$ and
            $\mathbf{60}\textbf{\%.}$
    } \\ \hline
\makecell[l]{Initial muscle radius dist.}
            & \makecell[l]{$\{
                    \text{Norm}(a,a), \text{Norm}(a, a^2), \text{Norm}(b,b), \text{Norm}(b, b^2), \mathbf{b,}$\\
                    $\quad \text{Unif}(a-a, a+a),\text{Unif}(a - a^2, a+a^2), \text{Unif}(b-b, b+b),$
                    \\
                    $\quad\text{Unif}(b - b^2, b+b^2)\}$,\\
            with 
            $a$ and $b$ corresponding to $O_C=57.5\%$ and $\mathbf{115}\textbf{\%}$.} \\ \hline
\end{tabular}
\centering
\label{table:design}
\endgroup
\end{table}

\begin{table}[t]
\caption{\textbf{Mold parameters.} The following parameters were used to 3D print the automatically-designed mold in which the robot was cast.}
\begin{tabular}{|l|l|}
\hline
\textbf{Variable}   & \textbf{Value} \\ \hline
Body width          & 5 cm           \\ \hline
Body height         & 3.5 cm         \\ \hline
Body thickness      & 1.4 cm         \\ \hline
Wall thickness      & 0.15 cm        \\ \hline
Actuation port size & 0.20 cm        \\ \hline
Dilation amount     & 1.3            \\ \hline
\end{tabular}
\centering
\label{table:molds}
\end{table}


\begin{thebibliography}{9}
\bibitem[1]{Sulsky1994} Sulsky, D., et al. A particle method for history-dependent materials. \textit{Computer Methods in Applied Mechanics and Engineering,} 118, 179-196 (1994)
\bibitem[2]{Griewank1989} Griewank, A. On automatic differentiation. Mathematical Programming: Recent Developments and Applications, 6, 83-107 (1989)
\bibitem[3]{Kingma2014} Kingma, D. Adam: A method for stochastic optimization. \textit{arXiv preprint arXiv:1412.6980} (2014).
\end{thebibliography}
\end{document}